# Generalizing Boolean Satisfiability I: Background and Survey of Existing Work

**Heidi E. Dixon**                                          DIXON@CIRL.UOREGON.EDU
*CIRL*
*Computer and Information Science*
*1269 University of Oregon*
*Eugene, OR 97403 USA*

**Matthew L. Ginsberg**                                     GINSBERG@OTSYS.COM
*On Time Systems, Inc.*
*1850 Millrace, Suite 1*
*Eugene, OR 97403 USA*

**Andrew J. Parkes**                                        PARKES@CIRL.UOREGON.EDU
*CIRL*
*1269 University of Oregon*
*Eugene, OR 97403 USA*

## Abstract

This is the first of three planned papers describing ZAP, a satisfiability engine that substantially generalizes existing tools while retaining the performance characteristics of modern high-performance solvers. The fundamental idea underlying ZAP is that many problems passed to such engines contain rich internal structure that is obscured by the Boolean representation used; our goal is to define a representation in which this structure is apparent and can easily be exploited to improve computational performance. This paper is a survey of the work underlying ZAP, and discusses previous attempts to improve the performance of the Davis-Putnam-Logemann-Loveland algorithm by exploiting the structure of the problem being solved. We examine existing ideas including extensions of the Boolean language to allow cardinality constraints, pseudo-Boolean representations, symmetry, and a limited form of quantification. While this paper is intended as a survey, our research results are contained in the two subsequent articles, with the theoretical structure of ZAP described in the second paper in this series, and ZAP's implementation described in the third.

## 1. Introduction

This is the first of a planned series of three papers describing ZAP, a satisfiability engine that substantially generalizes existing tools while retaining the performance characteristics of modern high-performance solvers such as zCHAFF (Moskewicz, Madigan, Zhao, Zhang, & Malik, 2001).[1] Many Boolean satisfiability problems incorporate a rich structure that reflects properties of the domain from which the problems themselves arise, and ZAP includes a representation language that allows this structure to be described and a proof engine that exploits the structure to improve performance. This first paper describes the work on which ZAP itself is based, and is intended to be a survey of existing results that attempt to use problem structure to improve the performance of satisfiability engines. The results we discuss include generalizations from Boolean satisfiability to include cardinality constraints, pseudo-Boolean representations, symmetry, and a limited form of quantification. The second paper in this series (Dixon et al., 2003b) describes the theoretical generalization that subsumes and extends these ideas, and the third paper (Dixon et al., 2003a) describes the ZAP system itself.

---

1. The second two papers have been published as technical reports (Dixon, Ginsberg, Luks, & Parkes, 2003b; Dixon, Ginsberg, Hofer, Luks, & Parkes, 2003a) but have not yet been peer reviewed.

Our intention is to review work on satisfiability from the introduction of the Davis-Putnam-Logemann-Loveland algorithm (Davis, Logemann, & Loveland, 1962) to the present day. Not all of this work (thankfully), but only that portion of the work that can be thought of as an attempt to improve the performance of systematic methods by exploiting the general structure of the problem in question.

We therefore include a description of recent work extending the language of Boolean satisfiability to include a restricted form of quantification (Ginsberg & Parkes, 2000) or pseudo-Boolean constraints (Barth, 1995, 1996; Chandru & Hooker, 1999; Dixon & Ginsberg, 2000; Hooker, 1988; Nemhauser & Wolsey, 1988); in each case, the representational extension corresponds to the existence of structure that is hidden by the Boolean axiomatization. We discuss at length the interplay between the desire to speed search by exploiting structure and the danger of slowing search by using unwieldy representations. Somewhat surprisingly, we will see that the most effective representational extensions appear to incur little or no overhead as a result of implementation concerns. It was this observation – that better representations can have better implementations – that led us to search for the general sort of structure that ZAP exploits.

We will also discuss the attempts that have been made to exploit the symmetrical structure of some satisfiability problems (Brown, Finkelstein, & Paul Walton Purdom, 1988; Crawford, 1992; Crawford, Ginsberg, Luks, & Roy, 1996; Joslin & Roy, 1997; Krishnamurthy, 1985; Puget, 1993). This work appears to have had only a modest impact on the development of satisfiability engines generally, and we explain why: Most authors (Crawford, 1992; Crawford et al., 1996; Joslin & Roy, 1997; Krishnamurthy, 1985; Puget, 1993) exploit only global symmetries, and such symmetries are vanishingly rare in naturally occurring problems. The methods that have been described for exploiting local or emergent symmetries (Brown et al., 1988; Szeider, 2003) incur unacceptable computational overhead at each node of the search. Our general arguments regarding the interplay between representation and search suggest that one should identify local symmetries when a problem is formulated, and then exploit those symmetries throughout the search.

We will *not* discuss heuristic search or any of the substantial literature relating to it. To our collective eye, just as a Boolean axiomatization can obscure the natural structure in a search problem, so can heuristics. We have argued elsewhere (Ginsberg & Geddis, 1991) that domain-dependent search control rules can never be more than a poor man's standin for general principles based on problem structure. Our selection of survey material reflects this bias.

We have remarked that this entry in the ZAP series is a survey paper; to the extent that there is a research contribution, it is in the overall (and, we believe, novel) focus we are taking. Our basic view is that the target of new work in this area should not be a specific representational extension such as a pseudo-Boolean or first-order encoding, but a direct exploitation of the underlying problem structure. This paper is original in describing existing work in this way, for the first time viewing the first-order and pseudo-Boolean extensions purely as structure-exploitation techniques. First-order and pseudo-Boolean representations are effective not because of their history of usefulness, but because – and to our mind *only* because – they allow the identification and capture of two particular types of structure inherent in many classes of problems. It is our hope that the reader views the material we are presenting here (and in the companion papers as well) in this light: Recent progress in Boolean satisfiability is best thought of in terms of structure exploitation. That is the perspective with which we approach this paper, and we hope that you, the reader, will come to share it.

But let us return to the Davis-Putnam-Logemann-Loveland procedure itself (Davis et al., 1962), which appears to have begun the body of work on the development of solvers that are sufficiently powerful to be used in practice on a wide range of problems. Descendants of the DPLL algorithm are now the solution method of choice on many such problems including microprocessor testing and verification (Biere, Clarke, Raimi, & Zhu, 1999; Copty, Fix, Giunchiglia, Kamhi, Tacchella, & Vardi, 2001; Velev & Bryant, 2001), and are competitive in other domains such as planning (Kautz & Selman, 1992).

We will return to the pure algorithmics of DPLL and its successors shortly, but let us begin by noting that in spite of impressive engineering successes on many difficult problems, there are many *easy* problems with which Boolean satisfiability engines struggle. These include problems involving parity, the well known "pigeonhole problem" (stating that you cannot put $n + 1$ pigeons into $n$ holes if each pigeon needs its own hole), and problems that are described most naturally using a first-order as opposed to a ground representation.

In all of these cases, there is structure to the problem being solved and this structure is lost when the ground encoding is built. While it is a testament to the power of Boolean methods that they can solve large and difficult problems without access to this underlying structure, it seems reasonable to expect that incorporating and using the structure would improve performance further. Our survey of techniques that improve DPLL will suggest that this is in fact the case, since all of the techniques that underpin the performance of modern Boolean satisfiability engines can be well understood in this way.

Before turning to the details of this analysis, however, let us flesh out our framework a bit more. If *any* computational procedure is to have its performance improved, there seem to be only three ways in which this can be done:

1. Replace the algorithm.

2. Reduce the time spent on a single iteration of the inner loop.

3. Reduce the number of times the inner loop must be executed.

The first approach is not our focus here; while there are many potential competitors to DPLL, none of them seems to outperform DPLL in practice.[2] Work on the second approach

---

2. Systematic alternatives include the original Davis-Putnam (1960) method, polynomial calculus solvers (Clegg, Edmonds, & Impagliazzo, 1996) based on Buchberger's (1965, 1985) Groebner basis algorithm, methods based on binary decision diagrams or BDDs (Bryant, 1986, 1992), and direct first-order methods has historically focused on reducing the amount of time spent in unit propagation, which does indeed appear to represent the inner loop of most DPLL implementations.

There are a variety of techniques available for reducing the number of calls to the inner loop, which we will divide as follows:

3a. Algorithmic improvements that do not require representational changes.

3b. Algorithmic improvements requiring representational changes.

Attempts to improve DPLL without changing the underlying Boolean representation have focused on (i) the development of a mechanism for retaining information developed in one portion of the search for subsequent use (caching or, more commonly, *learning*) and (ii) the development of search heuristics that can be expected to reduce the size of the space being examined. In all such work, however, using DPLL to analyze problems where no solution exists is equivalent to building a resolution proof of the unsatisfiability of the underlying theory, so that the number of inference steps is bounded by the number of inferences in the shortest such proof (Mitchell, 1998). These lengths are generally exponential in problem size in the worst case.

Work involving representation change can overcome this difficulty by reducing the length of the proof that DPLL or a similar algorithm is implicitly trying to construct. Representations are sought for which certain problems known to require proofs of exponential length in the Boolean case admit proofs of polynomial length after the representational shift is made. This leads to a hierarchy of representational choices, where one representation $r_1$ is said to polynomially simulate or *p-simulate* another $r_2$ if proofs using the representation $r_2$ can be converted to proofs using $r_1$ in polynomial time. In general, representations that lead to efficient proofs do so via more efficient encodings, so that a single axiom in the improved representation corresponds to exponentially many in the original. There are many excellent surveys of the proof complexity literature (Beame & Pitassi, 2001; Pitassi, 2002; Urquhart, 1995), and we will generally not repeat that material here.

Of course, it is not sufficient to simply improve the epistemological adequacy of a proof system; its heuristic adequacy must be maintained or improved as well (McCarthy, 1977). We therefore assume that any representation introduced for the purposes of navigating the $p$-simulation hierarchy also preserves the basic inference mechanism of DPLL (resolution) and maintains, and ideally builds upon, the improvements made in propagation performance (2) and learning (3a). It is in this context that our survey shall consider these representational changes.

The representations that we will consider are the following:

- Boolean axiomatizations. This is the original representation used in DPLL, and provides the basic setting in which progress in propagation, learning and branching has taken place.

---

such as those employed by OTTER (McCune & Wos, 1997). Nonsystematic methods include the WSAT family (Selman, Kautz, & Cohen, 1993), which received a great deal of attention in the 1990's and still appears to be the method of choice for randomly generated problems and some specific other sets of instances. In general, however, systematic algorithms with their roots in DPLL tend to outperform the alternatives.

- Cardinality constraints. If a disjunctive Boolean axiom states that at least one of the disjuncts is true, a cardinality constraint allows one to state that at least $n$ of the disjuncts are true for some integer $n$.

- Pseudo-Boolean constraints. Taken from operations research, a pseudo-Boolean constraint is of the form
$$\sum_i w_i l_i \geq k$$
where the $w_i$ are positive integral weights, the $l_i$ are literals, and $k$ is a positive integer. Cardinality constraints are the special case where $w_i = 1$ for all $i$; a Boolean constraint has $k = 1$ as well.

- Symmetric representations. Some problems (such as the pigeonhole problem) are highly symmetric, and it is possible to capture this symmetry directly in the axiomatization. A variety of authors have developed proof systems that exploit these symmetries to reduce proof size (Brown et al., 1988; Crawford, 1992; Crawford et al., 1996; Joslin & Roy, 1997; Krishnamurthy, 1985; Puget, 1993).

- Quantified representations. While there are many approaches to quantified satisfiability, we will focus only on those that do not change the underlying complexity of the problem being solved.[3] This requires that all domains of quantification be finite, and is the focus of a "propositional" restriction of first-order logic known as QPROP (Ginsberg & Parkes, 2000).

Given the arguments regarding heuristic adequacy generally, our goal in this survey is to complete the following table:

| | representational efficiency | p-simulation hierarchy | inference | unit propagation | learning |
|---|---|---|---|---|---|
| SAT cardinality pseudo-Boolean symmetry QPROP | | | | | |

The first column simply names the representational system in question. Then for each, we describe:

- Representational efficiency (3b): How many Boolean axioms can be captured in a single axiom in the given representation?

- p-simulation hierarchy (3b): Where is the representation relative to others in the p-simulation hierarchy?

- Inference: Is it possible to lift the basic DPLL inference mechanism of resolution to the new representation without incurring significant additional computational expense?

---

3. Problems involving quantified Boolean formulae, or QBF, are PSPACE-complete (Cadoli, Schaerf, Giovanardi, & Giovanardi, 2002) as opposed to the NP-complete problems considered by DPLL and its direct successors.

- Unit propagation (2): Can the techniques used to speed unit propagation be lifted to the new setting? Are new techniques available that are not available in the Boolean case?

- Learning (3a): Can existing learning techniques be lifted to the new setting? Are new techniques available that are not available in the Boolean case?

We cannot overstress the fact that no single column in our table is more important than the others. A reduction in proof length is of little practical value if there is no associated reduction in the amount of computation time required to find a proof. Speeding the time needed for a single inference is hardly useful if the number of inferences required grows exponentially.

The Boolean satisfiability community has pioneered many techniques used to reduce per-inference running time and to understand learning in a declarative setting. In spite of the fact that resolution and Boolean satisfiability are among the weakest inference systems in terms of representational efficiency and their position in the $p$-simulation hierarchy (Pudlak, 1997), almost none of the more powerful proof systems is in wide computational use. Finding proofs in these more sophisticated settings is difficult; even direct attempts to lift DPLL to a first-order setting (Baumgartner, 2000) seem fraught with complexity and an inability to exploit recent ideas that have led to substantial improvements in algorithmic performance. The most usable proof systems are often not the theoretically most powerful.

This paper is organized along the rows of the table we are trying to complete. Boolean techniques are described in the next section; we recount the demonstration that the pigeonhole problem is exponentially difficult in this setting (Haken, 1985). We go on in Section 3 to discuss cardinality constraints and pseudo-Boolean methods, showing that the earlier difficulties with the pigeonhole problem can be overcome using either method but that similar issues remain in other cases. Following the descriptions of implemented pseudo-Boolean reasoning systems (Barth, 1995; Dixon & Ginsberg, 2000), we show that the key computational ideas from the Boolean case continue to be applicable in a pseudo-Boolean setting.

Axiomatizations that attempt to exploit symmetry directly are discussed in Section 4. We draw a distinction between approaches that require the existence of global symmetries, which tend not to exist in practice, and those that use only local ones, which exist but are difficult to find as inference proceeds.

In Section 5, we discuss axiomatizations that are Boolean descriptions of problems that are more naturally represented using quantified axioms. We discuss the problems arising from the fact that the ground theories tend to be exponentially larger than their lifted counterparts, and show that working with the first-order axiomatization directly can lead to large improvements in the efficiency of the overall system (Ginsberg & Parkes, 2000).

Concluding remarks are contained in Section 6.

## 2. Boolean Satisfiability

**Definition 2.1** *A* variable *is simply a letter (e.g., a) that can be either true or false. A* literal *is either a variable or the negation of a variable. A* clause *is a disjunction of literals,* *and a* Boolean satisfiability problem (in conjunctive normal form), *or a* SAT *problem, is a conjunction of clauses.*

*A* solution *to a* SAT *problem C is an assignment of values to each of the letters so that every clause in C is satisfied.*

None of this should be new. Satisfiability of SAT instances is well-known to be NP-complete (Cook, 1971), and the language is a reasonably natural one for encoding real-world problems. As we remarked in the introduction, the classic algorithm for solving these problems is depth-first search augmented with an ability to set variables whose values are forced:

**Procedure 2.2 (Davis-Putnam-Logemann-Loveland)** *Given a* SAT *problem C and a partial assignment P of values to variables, to compute* DPLL$(C, P)$:

1  $P \leftarrow$ UNIT-PROPAGATE$(P)$
2  **if** $P$ contains a contradiction
3      **then return** FAILURE
4  **if** $P$ assigns a value to every variable
5      **then return** SUCCESS
6  $l \leftarrow$ a literal not assigned a value by $P$
7  **if** DPLL$(C, P \cup \{l = \texttt{true}\}) =$ SUCCESS
8      **then return** SUCCESS
9      **else return** DPLL$(C, P \cup \{l = \texttt{false}\})$

Variables are assigned values in two ways. In the first, unit propagation, the clause set is examined under the existing partial assignment and new consequential assignments are identified. Somewhat more specifically (see below), clauses are found that have no satisfied literals and exactly one unvalued literal. In each such clause, the unvalued literal is valued so as to satisfy the clause. This process is repeated until a contradiction is encountered, a solution is found, or no more clauses meet the necessary conditions. If the unit propagation function terminates without reaching a contradiction or finding a solution, then a variable is selected and assigned a value, and the procedure recurs.

In practice, the choice of branch literal is crucial to the performance of the algorithm. (Note that by choosing a branch literal $\neg l$ instead of $l$, we can also select the order in which values are tried for the underlying variable.) Relatively early work on DPLL focused on the selection of branch variables that produced as many unit propagations as possible, thus reducing the size of the residual problems that had to be solved recursively. As we will see in Section 2.3, however, more recent ideas appear to be more effective.

Missing from Procedure 2.2, however, is a description of the propagation process. Here it is:

**Procedure 2.3 (Unit propagation)** *To compute* UNIT-PROPAGATE$(P)$:

1  **while** no contradiction is found **and** there is a $c \in C$ that under $P$
      has no satisfied literals and exactly one unassigned literal
2          **do** $v \leftarrow$ the variable in $c$ unassigned by $P$
3              $P \leftarrow P \cup \{v = V : V$ is selected so that $c$ is satisfied$\}$
4  **return** $P$

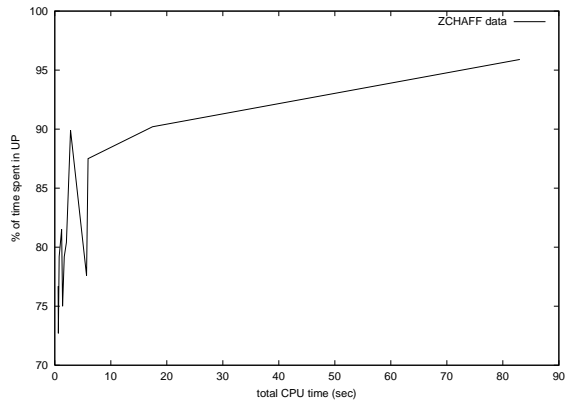

Figure 1: Fraction of CPU time spent in unit propagation

DPLL has a variety of well-known theoretical properties. It is sound and complete in that every candidate solution returned is a solution of the original problem, and such a solution will be returned if one exists (and failure eventually reported if no solution exists).

From a practical point of view, the running time of DPLL is obviously potentially exponential in the size of the problem, since at each iteration, possibly only a single variable is assigned a value before the routine is invoked recursively. In practice, of course, unit propagation can reduce the number of branch points substantially, but the running time remains exponential on difficult instances (Crawford & Auton, 1996).

We also point out that on difficult problem instances, most of the running time is necessarily spent exploring portions of the search space in which there are no solutions. After all, if $P$ is a partial assignment that can be extended to a solution, the algorithm will never backtrack away from $P$. (Indeed, it cannot and retain completeness, since $P$ may be the only variable assignment that extends to a full solution at this point.) Given that there can be no backtrack away from a satisfiable partial assignment, and that the number of backtracks is exponential in the problem size, it is clear that most of the time spent by the program is indeed in evaluating unsatisfiable regions of the search space.

## 2.1 Unit Propagation: The Inner Loop

When the DPLL algorithm 2.2 is implemented and run on practical problems, the bulk of the running time is spent in unit propagation. As an example, Figure 1 gives the amount of time spent by ZCHAFF on a variety of microprocessor test and verification examples made available by Velev (http://www.ece.cmu.edu/~mvelev).[4] As the problems become more difficult, an increasing fraction of the computational resources are devoted to unit propagation. For this reason, much early work on improving the performance of DPLL focused on improving the speed of unit propagation.

---

4. The examples used to generate the graph are those solved by ZCHAFF within 100 seconds using an Intel Pentium 4M running at 1.6 GHz. For those not solved within the 100 second limit, an average of 89.4% of the time was spent unit propagating.

Within the unit propagation procedure 2.3, the bulk of the time is spent identifying clauses that propagate; in other words, clauses that are not satisfied by the partial assignment and contain at most one unvalued literal:

**Observation 2.4** *Efficient implementations of* DPLL *typically spend the bulk of their effort searching for clauses satisfying the conditions required for unit propagation.*

Before we go on to examine the techniques that have been used to speed unit propagation in practice, let us remark that other implementations of SAT solvers have similar properties. Nonsystematic solvers such as WSAT (Selman et al., 1993), for example, spend the bulk of their time looking for clauses containing no satisfied or unvalued literals (or, equivalently, maintaining the data structures needed to make such search efficient). We can generalize Observation 2.4 to get:

**Observation 2.5** *Efficient implementations of* SAT *solvers typically spend the bulk of their effort searching for clauses satisfying specific syntactic conditions relative to a partial or complete truth assignment.*

While the focus of our survey is on systematic methods, we remark that because of the similarity of the techniques used in DPLL and in WSAT, techniques that speed the inner loop of one are likely to speed the inner loop of the other as well.

That said, let us describe the series of ideas that have been employed in speeding the process of identifying clauses leading to unit propagations:

1. After binding a variable $v$, examine each clause to determine whether or not it satisfies the conditions of Procedure 2.3.

2. Slightly more sophisticated is to restrict the search for a suitable clause to those clauses $c \in C$ that include $\neg v$ as one of the disjuncts (assuming that $v$ has been assigned the value true). After all, if $v$ appears in $c$, $c$ is satisfied after $v$ is set to true; if $v$ is not mentioned in $c$ there can be no change in $c$'s ability to unit propagate when $v$'s value is set.

3. When we set $v$ to true above, as we examine clauses containing $\neg v$, we have to walk each such clause to determine just which literals are satisfied (if any) and which are still unbound. It is more efficient to keep a record, for each clause $c$, of the number $s(c)$ of satisfied and the number $u(c)$ of unbound literals.

   In order to keep these counts current when we set a variable $v$ to true, we need to increment $s(c)$ and decrement $u(c)$ for each clause where $v$ appears, and to simply decrement $u(c)$ for each clause where $\neg v$ appears. If we backtrack and unset $v$, we need to reverse these adjustments.

   Compared to the previous approach, we need to examine four times as many clauses (those where $v$ appears with either sign, and both when $v$ is set and unset), but each examination takes constant time instead of time proportional to the clause length. If the average clause length is greater than four, this approach, due to Crawford and Auton (1996), will be more effective than its predecessor.

4. Currently, the most efficient scheme for searching for unit propagations is the *watched literals* family of implementations (Moskewicz et al., 2001; Zhang & Stickel, 2000). For each clause $c$, we identify two "watched" literals $l_1(c)$ and $l_2(c)$; the basic idea is that as long as these two literals are both either unvalued or satisfied, the clause cannot produce a unit propagation.

It is only when one of the watched literals is set to false that the clause must be examined in detail. If there is another unset (and unwatched) literal, we watch it. If there is a satisfied literal in the clause, we need do nothing. If there is no satisfied literal, the clause is ready to unit propagate.

If the average clause length is $l$, then when we set a variable $v$ (say to true), the probability is approximately $2/l$ that we will need to analyze a clause in which $\neg v$ appears, and the work involved is proportional to the length of the clause. So the expected amount of work involved is twice the number of clauses in which $\neg v$ appears, an improvement on the previous methods. In fact, the approach is somewhat more efficient than this cursory analysis suggests because the adjustment of the watched literals tends to favor watching those that are set deep in the search tree.

Before we move on to discuss learning in Boolean satisfiability, let us remark briefly on the so-called "pure literal" rule. To understand the rule itself, suppose that we have a theory $T$ and some partial variable assignment $P$. Suppose also that while a literal $q$ appears in some of the clauses in $T$ that are not yet satisfied by $P$, the negation $\neg q$ does not appear in any such clauses. Now while $q$ may not be a consequence of the partial assignment $P$, we can clearly set $q$ to true without removing any solutions from the portion of the search space that remains.

The pure literal rule is not generally included in implementations of either the DPLL or the unit propagation procedure because it is relatively expensive to work with. Counts of the number of unsatisfied clauses containing variables and their negations must be maintained at all times, and checked to see if a literal has become pure. In addition, we will see in Section 2.3 that many branching heuristics obviate the need for the pure literal rule to be employed.

## 2.2 Learning and Relevance

Let us now turn to the final column of our table, considering the progress that has been made in avoiding rework in Boolean satisfiability engines. The basic idea is to avoid situations where a conventional implementation will "thrash", solving the same subproblem many times in different contexts.

To understand the source of the difficulty, consider an example involving a SAT problem $C$ with variables $v_1, v_2, \ldots, v_{100}$, and suppose also that the subset of $C$ involving only variables $v_{50}, \ldots, v_{100}$ in fact implies that $v_{50}$ must be true.

Now imagine that we have constructed a partial solution that values the variables $v_1, \ldots, v_{49}$, and that we initially set $v_{50}$ to false. After some amount of backtracking, we realize that $v_{50}$ must be true. Unfortunately, if we subsequently change the value of one of the $v_i$'s for $i < 50$, we will "forget" that $v_{50}$ needs to be true and are in danger of setting it to false once again, followed by a repetition of the search that showed $v_{50}$ to be a consequence of $C$. Indeed, we are in danger of solving this subproblem not just twice, but once for each search node that we examine in the space of $v_i$'s for $i < 50$.

As we have indicated, the solution to this problem is to cache the fact that $v_{50}$ needs to be true; this information is generally saved in the form of a new clause called a *nogood*. In this case, we record the unary clause $v_{50}$. Modifying our original problem $C$ in this way will allow us to immediately prune any subproblem for which we have set $v_{50}$ to false. This technique was introduced by Stallman and Sussman (1977) in *dependency directed backtracking*.

Learning new constraints in this way can prevent thrashing. When a contradiction is encountered, the set of assignments that caused the contradiction is identified; we will call this set the *conflict set*. A new constraint can then be constructed that excludes the assignments in the conflict set. Adding this constraint to the problem will ensure that the faulty set of assignments is avoided in the future.

This description of learning is fairly syntactic; we can also give a more semantic description. Suppose that our partial assignment contains $\{a, \neg b, d, \neg e\}$ and that our problem contains the clauses

$$\neg a \lor b \lor c \lor \neg d \lor e \tag{1}$$

and

$$\neg c \lor \neg d. \tag{2}$$

The first clause unit propagates to allow us to conclude $c$; the second allows us to conclude $\neg c$. This contradiction causes us to backtrack, learning the nogood

$$\neg a \lor b \lor \neg d \lor e. \tag{3}$$

From a semantic point of view, the derived nogood (3) is simply the result of resolving the reason (1) for $\neg c$ with the reason (2) for $c$.

This is a general phenomenon. At any point, suppose that $v$ is a variable that has been set in the partial assignment $P$. If $v$'s value is the result of a branch choice, there is no associated reason. If $v$'s current value is the result of a unit propagation, however, we associate to $v$ as a reason the clause that produced the propagation. If $v$'s value is the result of a backtrack, that value must be the result of a contradiction identified for some subsequent variable $v'$ and we set the reason for $v$ to be the result of resolving the reasons for $v'$ and $\neg v'$. At any point, any variable whose value has been forced will have an associated reason, and these accumulated reasons will avoid the need to reexamine any particular portion of the search space.

Modifying DPLL to exploit the derived information requires that we include the derived clauses in the overall problem $C$, thus enabling new unit propagations and restricting subsequent search. But there is an implicit change as well.

In our earlier example, suppose that we have set the variables $v_1, \ldots, v_{49}$ in that order, and that we have learned the nogood

$$v_7 \lor \neg v_9 \tag{4}$$

(presumably in a situation where $v_7$ is false and $v_9$ is true). Now as long as $v_7$ remains false and $v_9$ remains true, unit propagation will fail immediately because (4) is unsatisfied. This will allow us to backtrack directly to $v_9$ in this example. This is the semantic justification for a technique known as *backjumping* (Gaschnig, 1979) because the search "jumps back" to a variable that is relevant to the problem at hand.[5]

While attractive in theory, however, this technique is difficult to apply in practice. The reason is that a new nogood is learned with every backtrack; since the number of backtracks is proportional to the running time of the program, an exponential number of nogoods can be learned. This can be expected both to overtax the memory available on the system where the algorithm is running, and to increase the time for each node expansion. As the number of clauses containing any particular variable $v$ grows without bound, the unit propagation procedure 2.3 will grind to a halt.

Addressing this problem was the primary focus of work on systematic SAT solvers during the 1990's. Since it was impractical to retain all of the nogoods that had been learned, some method needed to be found that would allow a polynomial number of nogoods to be retained while others were discarded. The hope was that a relatively small number of nogoods could still be used to prune the search space effectively.[6]

**Length-bounded Learning**   The first method used to bound the size of the set of learned clauses was to only retain clauses whose length is less than a given bound $k$ (Dechter, 1990; Frost & Dechter, 1994). In addition to providing a polynomial bound on the number of nogoods retained, this approach was felt likely to retain the most important learned clauses, since a clause of length $l$ will in general prune $\frac{1}{2^l}$ of the possible assignments of values to variables. Length-bounded learning draws from this observation the conclusion that short clauses should be retained in preference to long ones.

**Relevance-bounded Learning**   Unfortunately, length may not be the best indicator of the value of a particular learned clause. If we restrict our attention to any particular subproblem in the search, some short clauses may not be applicable at all, while other longer clauses may lead to significant pruning. As an example, consider a node defined by the partial assignment $P = \{a, b, c\}$ together with the two learned clauses:

$$a \lor b \lor e \tag{5}$$

$$\neg a \lor \neg b \lor \neg c \lor d \lor e \tag{6}$$

As long as the assignments in $P$ are retained, the clause (5) cannot be used for pruning because it is satisfied by $P$ itself. In fact, this clause will not be useful until we backtrack and change the values for both $a$ and for $b$. The clause (6) is more likely to be useful in the current subproblem, since it will lead to a unit propagation if either $d$ or $e$ is set to `false`. If the subproblem below the node given by $P$ is large, (6) may be used many times. Within the context of this subproblem, the longer clause is actually the more useful.

---

5. In this particular example, it is also possible to backtrack over $v_8$ as well, although no reason is recorded. The branch point for $v_8$ is removed from the search, and $v_9$ is set to false by unit propagation. The advantage of this is that there is now flexibility in either the choice of value for $v_8$ or the choice of branch variable itself. This idea is related to Baker's (1994) work on the difficulties associated with some backjumping schemes, and is employed in zChaff (Moskewicz et al., 2001).

6. Indeed, the final column of our table might better be named "forgetting" than "learning". Learning (everything) is easy; it's forgetting in a coherent way that's hard.

At some level, the appearance of $\neg a$, $\neg b$, and $\neg c$ in (6) shouldn't contribute to the effective length of the learned clause because that all of these literals are currently false. Length-bounded learning cannot make this distinction.

We see from this example that it is useful to retain clauses of arbitrary length provided that they are relevant to the current search context. If the context subsequently changes, we can remove some clauses to make room for new ones that are more suitable to the new context. This is what relevance-bounded learning does.

In relevance-bounded learning, the effective length of a clause is defined in terms of the current partial assignment. The *irrelevance* of a clause is defined as one less than the number of unvalued or satisfied literals it contains. For example, the clause (6) under the partial assignment $P$ has an irrelevance of 1. The idea of relevance-bounded learning originated in dynamic backtracking (Ginsberg, 1993), in which clauses were deleted if their irrelevance exceeded 1. This idea was generalized by Bayardo and Miranker (1996), who defined a general relevance bound and then deleted all clauses whose irrelevance exceeded that bound. This generalization is implemented in the RELSAT satisfiability engine (Bayardo & Schrag, 1997).

Like length-bounded learning, relevance-bounded learning retains all clauses of length less than the irrelevance bound $k$, since the irrelevance of a clause can never exceed its length. But the technique of relevance-bounded learning also allows the retention of longer clauses if they are applicable to the current portion of the search space. Such clauses are only removed when they are no longer relevant.

Returning to our example, if we backtrack from the original partial assignment with $\{a, b, c\}$ and find ourselves exploring $\{\neg a, \neg b, \neg c\}$, the short nogood (5) will be 0-irrelevant (since we can unit propagate to conclude $e$) and the long one (6) will be 4-irrelevant. Using a relevance bound of 3 or less, this nogood would be discarded.

The condition that nogoods are only retained until their irrelevance exceeds a bound $k$ is sufficient to ensure that only a polynomial number of nogoods are retained at any point.[7] Experimental results (Bayardo & Miranker, 1996) show that relevance-bounded learning is more effective than its length-bounded counterpart and, even with a relevance bound of 1, the results are comparable to those of learning without restriction.[8]

**Hybrid Approaches**   Finally, we note that some solvers employ a hybrid approach. The CHAFF algorithm (Moskewicz et al., 2001) uses both a relevance bound and a (larger) length bound. Clauses must meet both the relevance bound and the length bound to be retained.

Yet another approach is taken by the BERKMIN algorithm (Goldberg & Novikov, 2002). Here, the set of nogoods is partitioned into two separate groups based on how recently the nogoods were acquired; $\frac{15}{16}$ of the nogoods are kept in a "recent" group and the remaining $\frac{1}{16}$ in an "old" group. A relatively large length bound is used to cull the recently acquired nogoods while a smaller length bound is used to aggressively cull the smaller group of older nogoods. We are not aware of any studies comparing these hybrid approaches to pure length-bounded or relevance-bounded methods.

---

7. Although accepted wisdom, we know of no proof in the literature of this result; Bayardo and Miranker's (1996) proof appears to assume that the order in which branch variables are chosen is fixed. We present a general proof in the next paper in this series (Dixon et al., 2003b).

8. Results such as these necessarily become somewhat suspect as algorithmic methods mature; an unfortunate consequence of the extremely rapid progress in satisfiability engines over recent years is the lack of careful experimental work evaluating the host of new ideas that have been developed.

## 2.3 Branching Heuristics

Let us now turn to an examination of the progress that has been made in the development of effective heuristics for the selection of branch variables in DPLL. As discussed in the introduction, we focus not on specific heuristics that work in selected domains, but on general ideas that attempt to exploit the structure of the axiomatization directly.

Prior to the development of successful learning techniques, branching heuristics were the primary method used to reduce the size of the search space. It seems likely, therefore, that the role of branching heuristics is likely to change significantly for algorithms that prune the search space using learning. While the heuristics used in zCHAFF appear to be a first step in this direction, very little is known about this new role.

Initially, however, the primary criterion for the selection of a branch variable was to pick one that would enable a cascade of unit propagations; the result of such a cascade is a smaller and more tractable subproblem.[9]

The first rule based on this idea is called the MOMS rule, which branches on the variable that has Maximum Occurrences in clauses of Minimum Size (Crawford & Auton, 1996; Dubois, Andre, Boufkhad, & Carlier, 1993; Hooker & Vinay, 1995; Jeroslow & Wang, 1990; Pretolani, 1993). MOMS provides a rough but easily computed approximation to the number of unit propagations that a particular variable assignment might cause.

Alternatively, one can use a "unit propagation rule" (Crawford & Auton, 1996; Freeman, 1995), and compute the exact number of propagations that would be caused by a branching choice. Given a branching candidate $v_i$, the variable is separately fixed to true and to false and the unit propagation procedure is executed for each choice. The precise number of unit propagations caused is then used to evaluate possible branching choices. Unlike the MOMS heuristic, this rule is obviously exact in its attempt to judge the number of unit propagations caused by a potential variable assignment. Unfortunately, it is also considerably more expensive to compute because of the expense of unit propagation itself. This led to the adoption of composite approaches (Li & Anbulagan, 1997) where MOMS is used to identify a small number of branching candidates, each of which is then evaluated exactly using the more expensive unit propagation heuristic. On randomly generated problems, the composite technique outperforms either heuristic in isolation.

Another strategy is to branch on variables that are likely to be *backbone* variables (Dubois & Dequen, 2001). A *backbone* literal (also often referred to as a *unary prime implicate* of a theory) is one that must be true in all solutions to a given problem. Given a problem $C$ and a partial assignment $P$, the backbone heuristic attempts to branch on variables that are backbones of the subset of those clauses in $C$ that are satisfied by $P$; the likelihood that any particular variable is a backbone literal is approximated by counting the appearances of that literal in the satisfied clauses in $C$. This heuristic outperforms those discussed in the previous paragraphs.

---

9. This idea tends to obviate the need for use of the pure literal rule, as well. If $p$ is a pure literal, there is no particular reason to hurry to set $p$ to true; the key thing is to avoid setting it to false. But if $p$ is pure, $\neg p$ cannot generate any unit propagations, so $p$ will tend not to be selected as a branch variable. Pure literals can obviously never be set false by unit propagation, so heuristics based on unit propagation counts tend to achieve most of the advantages of the pure literal rule without incurring the associated computational costs.

The heuristics described thus far were developed when the community's research emphasis was focused on the solution of randomly generated satisfiability problems. The development of bounded learning methods enabled solvers to address the issue of thrashing, and caused a natural shift in focus toward more structured, realistic problems. There are no formal studies comparing the previously discussed heuristics on structured problems, and the value of the studies that do exist is reduced because all of the implementations were of DPLL in isolation and without the learning techniques that have since proved so important. Branching techniques and learning are deeply related, and the addition of learning to a DPLL implementation will have a significant effect on the effectiveness of any of these branching strategies. As new clauses are learned, the number of unit propagations an assignment will cause can be expected to vary; the reverse is also true in that the choice of branch variable can affect which clauses the algorithm learns. A formal comparison of branching techniques' performance on structured problems and in the presence of learning would be extremely useful.

Branching heuristics that are designed to function well in the context of a learning algorithm generally try to branch on variables about which things have been learned recently. This tends to allow the implementation to keep "making progress" on a single section of the search space as opposed to moving from one area to another; an additional benefit is that existing nogoods tend to remain relevant, avoiding the inefficiencies associated with losing the information present in nogoods that become irrelevant and are deleted. In zCHAFF, for example, a count is maintained of the number of times each literal occurs in the theory being solved. When a new clause is added, the count associated with each literal in the clause is incremented. The branch heuristic then selects a variable that appears in as many clauses as possible. By periodically dividing all of the counts by a constant factor, a bias is introduced toward branching on variables that appear in recently learned clauses. Like the MOMS rule, this rule is inexpensive to calculate.

The heuristic used in BERKMIN (Goldberg & Novikov, 2002) builds on this idea but responds more dynamically to recently learned clauses. The BERKMIN heuristic prefers to branch on variables that are unvalued in the most recently learned clause that is not yet satisfied, with a zCHAFF-like heuristic used to break ties.

All told, there are many competing branching heuristics for satisfiability solvers, and there is still much to be done in evaluating their relative effectiveness. The most interesting experiments will be done using implementations that learn, and on realistic, structured problems as opposed to randomly generated ones.

## 2.4 Proof Complexity

We have already commented briefly on the fact that proof systems can be evaluated based on provable bounds on the proofs of certain classes of formulae, or by the development of polynomial transformations from proofs in one system into proofs in another.

With regard to the first metric, there are at least three classes of problems known to be exponentially difficult for conventional resolution-based provers (including any DPLL implementation):[10]

1. Pigeonhole problems (Haken, 1985)

2. Parity problems (Tseitin, 1970)

3. Clique coloring problems (Bonet, Pitassi, & Raz, 1997; Kraj'icek, 1997; Pudlak, 1997)

Before turning to a discussion of these problems specifically, however, let us point out that there are many proof systems that are known to be more powerful than any of the ones we discuss in this paper. From our perspective, the most interesting is *extended resolution* and involves the introduction of new variables that correspond to arbitrary logical expressions built up out of the original variables in the theory.

Since such logical expressions can always be built up term-by-term, it suffices to allow the introduction of new variables corresponding to pairwise combinations of existing ones; since disjunction can be replaced with conjunction using de Morgan's laws, it suffices to introduce new variables of the form

$$w \equiv x \wedge y \tag{7}$$

for literals $x$ and $y$. Writing (7) in disjunctive normal form, we get:

**Definition 2.6 (Tseitin, 1970)** *An extended resolution proof for a theory $T$ is one where $T$ is first augmented by a collection of groups of axioms, each group of the form*

$$
\begin{aligned}
\neg x \vee \neg y \vee w \\
x \vee \neg w \\
y \vee \neg w
\end{aligned}
\tag{8}
$$

*where $x$ and $y$ are literals in the (possibly already extended) theory $T$ and $w$ is a new variable. Following this, derivation proceeds using conventional resolution on the augmented theory.*

There is no proof system known to be stronger than extended resolution; in fact, there is no class of problems for which there are known to be no polynomially sized proofs in extended resolution.

As we have stressed, however, the fact that a proof system is strong does not mean that it works well in practice. We know of *no* implementation of extended resolution for the simple reason that virtually nothing is known about how to select new variables so as to shorten proof length.

Understanding *why* the introduction of these new variables can reduce proof length is considerably simpler. As an example, suppose that during a resolution proof, we have managed to derive the nogood $a \vee x$, and that we have also derived $a \vee y$. In order to complete the proof, we need to perform lengthy – but identical – analyses of each of these nogoods, eventually deriving simply $x$ from the first and $y$ from the second (and then resolving against $\neg x \vee \neg y$, for example).

---

10. There are other hard problems as well, such as Haken's (1995) *broken mosquito screen* problem. The three examples quoted here are sufficient for our purposes.

If we could replace the pair of nogoods $a \lor x$ and $a \lor y$ with the single nogood $a \lor w$ using (8), the two proofs from $a \lor x$ and $a \lor y$ could be collapsed into a single proof, potentially halving the size of the proof in its entirety. Introducing still more variables can repeat the effect, resulting in exponential reductions in proof size.

Another way to look at this is as an improvement in expressivity. There is simply no way to write $(a \lor x) \land (a \lor y)$ or the equivalent $a \lor (x \land y)$ as a single Boolean axiom in disjunctive normal form. The power of extended resolution rests on the fact that subexpression substitution makes it possible to capture expressions such as $a \lor (x \land y)$ in a single axiom.

None of the proof systems being considered in this survey is as powerful as extended resolution, and we will therefore evaluate them based on their performance on the three problems mentioned at the beginning of this section. Let us therefore describe each of those problems in some detail.

### 2.4.1 Pigeonhole problems

The pigeonhole problem involves showing that it is impossible to put $n + 1$ pigeons into $n$ holes if each pigeon must go into a distinct hole. If we write $p_{ij}$ for the fact that pigeon $i$ is in hole $j$, then a straightforward axiomatization says that every pigeon must be in at least one hole:

$$p_{i1} \lor p_{i2} \lor \cdots \lor p_{in} \text{ for } i = 1, \ldots, n + 1 \tag{9}$$

and that no two pigeons can be in the same hole:

$$\neg p_{ik} \lor \neg p_{jk} \text{ for } 1 \leq i < j \leq n + 1 \text{ and } k = 1, \ldots, n \tag{10}$$

Note that there are in all $\Theta(n^3)$ axioms of the form (10).

It is well known that there is no polynomial-sized proof of the unsatisfiability of the axioms (9)–(10) (Haken, 1985). The proof is technical, but the essential reason is that the pigeonhole problem is "all about" counting. At some point, proving that you can't put $n + 1$ pigeons into $n$ holes requires saying that you can't put $n$ pigeons into the last $n - 1$ holes, thus $n - 1$ pigeons into the last $n - 2$ holes, and so on. Saying this in the language of SAT is awkward, and it is possible to show that no proof of the pigeonhole problem can be completed without, at some point, working with extremely long individual clauses. Once again, we see the connection to expressive efficiency; for readers interested in additional details, Pitassi's (2002) explanation is reasonably accessible.

### 2.4.2 Parity problems

By a *parity problem*, we will mean a collection of axioms specifying the parity of sets of inputs. So we will write, for example,

$$x_1 \oplus \cdots \oplus x_n = 1 \tag{11}$$

to indicates that an odd number of the $x_i$ are true; a right hand side of zero would indicate that an even number were true. The $\oplus$ here indicates exclusive or.

Reduction of (11) to a collection of Boolean axioms is best described by an example. The parity constraint $x \oplus y \oplus z = 1$ is equivalent to

$$x \vee y \vee z$$
$$x \vee \neg y \vee \neg z$$
$$\neg x \vee y \vee \neg z$$
$$\neg x \vee \neg y \vee z$$

In general, the number of Boolean axioms needed is exponential in the length of the parity clause (11), but for clauses of a fixed length, the number of axioms is obviously fixed as well.

For the proof complexity result of interest, suppose that $G$ is a graph, where each node in $G$ will correspond to a clause and each edge to a literal. We label the edges with distinct literals, and label each node of the graph with a zero or a one. Now if $n$ is a node of the graph that is labeled with a value $v_n$ and the edges $e_{1n}, \ldots, e_{i(n),n}$ incident on $n$ are labeled with literals $l_{1n}, \ldots, l_{i(n),n}$, we add to our theory the Boolean version of the clause

$$l_{1n} \oplus \cdots \oplus l_{i(n),n} = v_n \tag{12}$$

Since every edge connects two nodes, every literal in the theory appears exactly twice in axioms of the form (12). Adding all of these constraints therefore produces a value that is equivalent to zero mod 2 and must be equal to $\sum_n v_n$ as well. If $\sum_n v_n$ is odd, the theory is unsatisfiable. Tseitin's (1970) principal result is to show that this unsatisfiability cannot in general be proven in a number of resolution steps polynomial in the size of the Boolean encoding.

### 2.4.3 CLIQUE COLORING PROBLEMS

The last examples we will consider are known as "clique coloring problems." These are derivatives of pigeonhole problems where the exact nature of the pigeonhole problem is obscured. Somewhat more specifically, the problems indicate that a graph includes a clique of $n + 1$ nodes (where every node in the clique is connected to every other), and that the graph must be colored in $n$ colors. If the graph itself is known to be a clique, the problem is equivalent to the pigeonhole problem. But if we know only that the clique can be embedded into the graph, the problem is more difficult.

In the axiomatization, we use $e_{ij}$ to describe the edges of the graph, $c_{ij}$ to describe the coloring of the graph, and $q_{ij}$ to describe the embedding of the cliQue into the graph. The graph has $m$ nodes, the clique is of size $n + 1$, and there are $n$ colors available. So the axiomatization is:

$$\neg e_{ij} \vee \neg c_{il} \vee \neg c_{jl} \qquad \text{for } 1 \leq i < j \leq m, \, l = 1, \ldots, n \tag{13}$$

$$c_{i1} \vee \cdots \vee c_{in} \qquad \text{for } i = 1, \ldots, m \tag{14}$$

$$q_{i1} \vee \cdots \vee q_{im} \qquad \text{for } i = 1, \ldots, n + 1 \tag{15}$$

$$\neg q_{ij} \vee \neg q_{kj} \qquad \text{for } 1 \leq i < k \leq n + 1, \, j = 1, \ldots, m \tag{16}$$

$$e_{ij} \vee \neg q_{ki} \vee \neg q_{lj} \qquad \text{for } 1 \leq i < j \leq m, \, 1 \leq k \neq l \leq n + 1 \tag{17}$$

Here $e_{ij}$ means that there is an edge between graph nodes $i$ and $j$, $c_{ij}$ means that graph node $i$ is colored with the $j$th color, and $q_{ij}$ means that the $i$th element of the clique is mapped to graph node $j$. Thus the first axiom (13) says that two of the $m$ nodes in the graph cannot be the same color (of the $n$ colors available) if they are connected by an edge. (14) says that every graph node has a color. (15) says that every element of the clique appears in the graph, and (16) says that no two elements of the clique map to the same node in the graph. Finally, (17) says that the clique is indeed a clique – no two clique elements can map to disconnected nodes in the graph.

Since there is no polynomially sized resolution proof of the pigeonhole problem in Boolean satisfiability, there is obviously no polynomially sized proof of the clique coloring problems, either. But as we shall see, clique coloring problems can in some cases be used to distinguish among elements of the proof complexity hierarchy.

## 2.5 Boolean Summary

We summarize the results of this section by completing the first row of our table as follows:

| | rep. eff. | p-simulation hierarchy | inference | unit propagation | learning |
|---|---|---|---|---|---|
| **SAT** **cardinality** **PB** **symmetry** **QPROP** | 1 | EEE | resolution | watched literals | relevance |

The entries are really just an informal shorthand:

- **Representational efficiency:** Boolean satisfiability is the benchmark against which other languages will be measured; we give here the relative savings to be had by changing representation.

- **p-simulation hierarchy:** We give the proof complexity for the three problem classes discussed in Section 2.4. For Boolean satisfiability, all of the problems require proofs of exponential length.

- **Inference:** The basic inference mechanism used by DPLL is resolution.

- **Propagation:** Watched literals lead to the most efficient implementation.

- **Learning:** Relevance-based learning appears to be more effective than other poly-sized methods.

## 3. Pseudo-Boolean and Cardinality Constraints

The entries in the previous table summarize the fact that Boolean satisfiability is a weak method that admits efficient implementations. The representation is relatively inefficient, and none of our canonical problems can be solved in polynomial time. Some of these difficulties, at least, can be overcome via a representational shift.

To understand the shift, note that we can write an axiom such as

$$x \vee y \vee \neg z$$

as

$$x + y + \overline{z} \geq 1 \tag{18}$$

where we are now thinking of $x, y$ and $z$ as variables with value either 0 (false) or 1 (true) and have written $\overline{z}$ for $1 - z$ or, equivalently, $\neg z$. If $v$ is a variable, we will continue to refer to $\overline{v}$ as the negation of $v$.

All of the familiar logical operations have obvious analogs in this notation. If, for example, we want to resolve

$$a \vee \neg b \vee c$$

with

$$b \vee \neg d$$

to get $a \vee c \vee \neg d$, we simply add

$$a + \overline{b} + c \geq 1$$

to

$$b + \overline{d} \geq 1$$

and simplify using the identity $b + \overline{b} = 1$ to get

$$a + c + \overline{d} \geq 1$$

as required.

What's nice about this notation is that it extends easily to more general descriptions. If the general form of a disjunction $\vee l_i$ of literals is $\sum l_i \geq 1$ as in (18), we can drop the requirement that the right-hand side be 1:

**Definition 3.1** *A* cardinality constraint *or* extended clause *is a constraint of the form*

$$\sum_i l_i \geq k$$

*where $k$ is an integer and each of the $l_i$ is required to have value 0 or 1.*

The cardinality constraint simply says that at least $k$ of the literals in question are true.

**Proposition 3.2 (Cook, Coullard and Turan (1987))** *There is an unsatisfiability proof of polynomial length of the pigeonhole problem using cardinality constraints.*

**Proof.** Cook et al. (1987) give a derivation in $o(n^3)$ steps; we have presented an $o(n^2)$ derivation elsewhere (Dixon & Ginsberg, 2000). ◻

Of course, the fact that this extension to the SAT language allows us to find polynomial-length derivations of pigeonhole problem does not necessarily show that the change will have computational value; we need to examine the other columns of the table as well. In the remainder of this section, we will show this and will go further, describing new computational techniques that can only be applied in the broader setting that we are now considering. Experimental results are also presented. But let us begin by examining the first column of the table:

**Proposition 3.3 (Benhamou, Sais, & Siegel, 1994)** *The cardinality constraint*

$$x_1 + \cdots + x_m \geq k \tag{19}$$

*is logically equivalent to the set of $\binom{m}{k-1}$ axioms*

$$x_{i_1} + \cdots + x_{i_{m-k+1}} \geq 1 \tag{20}$$

*for every set of $m - k + 1$ distinct variables $\{x_{i_1}, \ldots, x_{i_{m-k+1}}\}$. Furthermore, there is no more compact Boolean encoding of (19).*

**Proof.** We first show that (19) implies (20). To see this, suppose that we have a set $S$ of $m - k + 1$ variables. Suppose also that $T$ is the set of $x_i$'s that are true, so that $T$ is of size at least $k$. Since there are only $m$ variables, $S \cap T \neq \emptyset$ and at least one $x_i \in S$ must be true.

To see that (20) implies (19), suppose that (20) is true for all appropriate sets of $x_i$'s. Now if (19) were false, the set of false $x_i$'s would be of size at least $m - k + 1$, so that some instance of (20) would be unsatisfied.

To see that there is no more efficient encoding, first note that if (19) implies a Boolean axiom

$$x_1 \vee \cdots \vee x_k \vee \neg x_{k+1} \vee \cdots \vee \neg x_m$$

then it must also imply

$$x_1 \vee \cdots \vee x_k$$

since we can always change an $x_i$ from false to true without reducing the satisfiability of (19).

Next, note that no axiom of length less than $m - k + 1$ is a consequence of (19), since any such axiom can be falsified while satisfying (19) by setting every unmentioned variable to true and the rest to false.

Finally, suppose that we leave out a single instance $i$ of (20) but include all of the others as well as every clause of length greater than $m - k + 1$. By setting the variables in $i$ to false and every other variable to true, all of the given clauses will be satisfied but (19) will not be. It follows that any Boolean equivalent of (19) must include at least the $\binom{m}{k-1}$ instances of (20). □

It follows from Proposition 3.3 that provided that no new variables are introduced, cardinality constraints can be exponentially more efficient than their Boolean counterparts.

Before discussing the other columns in the table, let us consider further extending our representation to include what are known as *pseudo*-Boolean constraints:

**Definition 3.4** *A* pseudo-Boolean constraint *is an axiom of the form*

$$\sum_i w_i l_i \geq k \tag{21}$$

*where each $w_i$ and $k$ is a positive integer and each of the $l_i$ is required to have value 0 or 1.*

Pseudo-Boolean representations typically allow both linear inequalities and linear equalities over Boolean variables. Linear equalities can easily be translated into a pair of inequalities of the form in the definition; we prefer the inequality-based description (Barth, 1996; Chandru & Hooker, 1999, also known as pseudo-Boolean *normal form*) because of the better analogy with Boolean satisfiability and because unit propagation becomes unmanageable if equality constraints are considered. Indeed, simply determining if an equality clause is satisfiable subsumes subset-sum and is therefore (weakly) NP-complete (Garey & Johnson, 1979).

Compare (21) with Definition 3.1; the $w_i$ are the weights attached to various literals. The pseudo-Boolean language is somewhat more flexible still, allowing us to say (for example)

$$2a + b + \overline{c} \geq 2$$

indicating that either $a$ is true or both $b$ and $\neg c$ are (equivalent to the crucial representational efficiency obtained in extended resolution). As we will see shortly, it is natural to make this further extension because the result of resolving two cardinality constraints can be most naturally written in this form.

## 3.1 Unit Propagation

Let us begin by discussing propagation techniques in a cardinality or pseudo-Boolean setting.[11]

A pseudo-Boolean version of unit propagation was first presented by Barth (1996) and is described in a number of papers (Aloul, Ramani, Markov, & Sakallah, 2002; Dixon & Ginsberg, 2000). In the Boolean case, we can describe a clause as unit if it contains no satisfied literals and at most one unvalued one. To generalize this to the pseudo-Boolean setting, we make the following definition, where we view a partial assignment $P$ simply as the set of literals that it values to true:

**Definition 3.5** *Let $\sum_i w_i l_i \geq k$ be a pseudo-Boolean clause, which we will denote by $c$. Now suppose that $P$ is a partial assignment of values to variables. We will say that the current value of $c$ under $P$ is given by*

$$\mathtt{curr}(c, P) = \sum_{\{i \mid l_i \in P\}} w_i - k$$

*If no ambiguity is possible, we will write simply $\mathtt{curr}(c)$ instead of $\mathtt{curr}(c, P)$. In other words, $\mathtt{curr}(c)$ is the sum of the weights of literals that are already satisfied by $P$, reduced by the required total weight $k$.*

*In a similar way, we will say that the possible value of $c$ under $P$ is given by*

$$\mathtt{poss}(c, P) = \sum_{\{i \mid \neg l_i \notin P\}} w_i - k$$

---

11. As we have remarked, our table is designed to reflect the issues involved in lifting DPLL to a more expressive representation. Extending a nonsystematic search technique such as WSAT to a pseudo-Boolean setting has been discussed by Walser (1997) and Prestwich (2002).

*If no ambiguity is possible, we will write simply* $\mathtt{poss}(c)$ *instead of* $\mathtt{poss}(c, P)$. *In other words,* $\mathtt{poss}(c)$ *is the sum of the weights of literals that are either already satisfied or not valued by* $P$, *reduced by the required total weight* $k$.[12]

**Definition 3.6** *Let $c$ be a clause, and $P$ a partial assignment. We will say that $c$ is* unit *if there is a variable $v$ not appearing in $P$ such that either $P \cup \{v\}$ or $P \cup \{\neg v\}$ cannot be extended to an assignment that satisfies $c$.*

In this situation, the variable $v$ is forced to take a value that will help satisfy the clause. This creates a new consequential assignment. Note that if $c$ is already unsatisfiable, we can meet the conditions of the definition by choosing $v$ to be any variable not assigned a value by $P$. Note also that in the pseudo-Boolean case, a clause may actually contain more than one variable that is forced to a specific value. It should be clear that in the Boolean case, this definition duplicates the conditions of the original unit propagation procedure 2.3.

**Lemma 3.7** *A partial assignment $P$ can be extended in a way that satisfies a clause $c$ if and only if $\mathtt{poss}(c, P) \geq 0$.*

**Proof.** Assume first that $\mathtt{poss}(c, P) \geq 0$, and suppose that we value every remaining variable in a way that helps to satisfy $c$. Having done so, every literal in $c$ that is not currently made false by $P$ will be true, and the resulting value of $c$ will be

$$\sum_i w_i l_i = \sum_{\{i | \neg l_i \notin P\}} w_i = \mathtt{poss}(c, P) + k \geq k$$

so that $c$ becomes satisfied.

Conversely, suppose that $\mathtt{poss}(c, P) < 0$. Now the best we can do is still to value the unvalued literals favorably, so that the value of $c$ becomes

$$\sum_i w_i l_i = \sum_{\{i | \neg l_i \notin P\}} w_i = \mathtt{poss}(c, P) + k < k$$

and $c$ is unsatisfiable. □

**Proposition 3.8** *A clause $c$ containing at least one unvalued literal is unit if and only if $c$ contains an unvalued literal $l_i$ with weight $w_i > \mathtt{poss}(c)$.*

**Proof.** If there is a literal with weight $w_i > \mathtt{poss}(c)$, setting that literal to false will reduce $\mathtt{poss}(c)$ by $w_i$, making it negative and thus making the $c$ unsatisfiable. Conversely, if there is no such literal, then $\mathtt{poss}(c)$ will remain positive after any single unvalued literal is set, so that $c$ remains satisfiable and is therefore not unit. □

Given the above result, there is little impact on the time needed to find unit clauses. We need simply keep the literals in each clause sorted by weight and maintain, for each clause, the value of $\mathtt{poss}$ and the weight of the largest unvalued literal. If we value a literal with different weight, we can apply the test in Proposition 3.8 directly. If we value a literal of the given weight, a short walk along the clause will allow us to identify the new unvalued literal of maximum weight, so that the proposition continues to apply.

---

12. Chai and Kuehlmann (2003) refer to $\mathtt{poss}$ as *slack*.

**Watched literals**   Generalizing the idea of watched literals is no more difficult. We make the following definition:

**Definition 3.9** *Let $c$ be a clause. A watching set for $c$ is any set $S$ of variables with the property that $c$ cannot be unit as long as all of the variables in $S$ are either unvalued or satisfied.*

**Proposition 3.10** *Given a clause $c$ of the form $\sum_i w_i l_i \geq k$, let $S$ be any set of variables. Then $S$ is a watching set for $c$ if and only if*

$$\sum_i w_i - \max_i w_i \geq k \qquad (22)$$

*where the sum and maximum are taken over literals involving variables in $S$.*

**Proof.** Suppose that all of the variables in $S$ are unvalued or satisfied. Now let $l_j$ be any unvalued literal in $c$. If $l_j \notin S$, then $\mathtt{poss}(c) \geq w_j + \sum_i w_i - k$ and thus $\mathtt{poss}(c) \geq w_j$ since $\sum_i w_i \geq \sum_i w_i - \max_i w_i \geq k$. If, on the other hand, $l_j \in S$, then

$$\mathtt{poss}(c) \geq \sum_i w_i - k$$

and

$$\sum_i w_i - w_j \geq \sum_i w_i - \max_i w_i \geq k$$

Combining these, we get

$$\mathtt{poss}(c) \geq w_j$$

Either way, we cannot have $\mathtt{poss}(c) < w_j$ and Proposition 3.8 therefore implies that $c$ cannot be unit. It follows that $S$ is a watching set.

The converse is simpler. If $\sum_i w_i - \max_i w_i < k$, value every literal outside of $S$ so as to make $c$ false. Now $\mathtt{poss}(c) = \sum_i w_i - k$, so if $l_j$ is the literal in $S$ with greatest weight, the associated weight $w_j$ satisfies $w_j > \mathtt{poss}(c)$ and $c$ is unit. Thus $S$ cannot be a watching set. $\square$

This generalizes the definition from the Boolean case, a fact made even more obvious by:

**Corollary 3.11** *Given a cardinality constraint $c$ requiring at least $k$ of the associated literals to be true, $S$ is a watching set for $c$ if and only if it includes at least $k + 1$ literals in $c$.*

**Proof.** The expression (22) becomes

$$\sum_i 1 - \max_i 1 \geq k$$

or

$$|S| - 1 \geq k. \qquad \square$$

### 3.2 Inference and Resolution

As with unit propagation, resolution also lifts fairly easily to a pseudo-Boolean setting. The general computation is as follows:

**Proposition 3.12** *Suppose that we have two clauses $c$ and $c'$, with $c$ given by*

$$\sum_i w_i l_i + wl \geq k \tag{23}$$

*and $c'$ given by*

$$\sum_i w_i' l_i' + w\overline{l} \geq k' \tag{24}$$

*Then it is legitimate to conclude*

$$\sum_i w' w_i l_i + \sum_i w w_i' l_i' \geq w'k + wk' - ww' \tag{25}$$

**Proof.** This is immediate. Multiply (23) by $w'$, multiply (24) by $w$, add and simplify using $l + \overline{l} = 1$. ☐

If all of the weights, $k$ and $k'$ are 1, this generalizes conventional resolution provided that the sets of nonresolving literals in $c$ and $c'$ are disjoint. To deal with the case where there is overlap between the set of $l_i$ and the set of $l_i'$, we need:

**Lemma 3.13** *Suppose that $c$ is clause $\sum_i w_i l_i \geq k$. Then $c$ is equivalent to $\sum_i w_i' l_i \geq k$, where the $w_i'$ are given by:*

$$w_i'(j) = \begin{cases} w_i, & \text{if } w_i < k; \\ k, & \text{otherwise.} \end{cases}$$

**Proof.** If $l_j$ is a literal with $w_j \geq k$, then both $c$ and the rewrite are true if $l_j$ is satisfied. If $l_j = 0$, then $c$ and the rewrite are equivalent. ☐

In other words, we can reduce any coefficient that is greater than what is required to satisfy the clause in its entirety, for example rewriting

$$3x + y + \overline{z} \geq 2$$

as

$$2x + y + \overline{z} \geq 2$$

because either is equivalent to $x \vee (y \wedge \neg z)$.

**Proposition 3.14 (Cook et al., 1987; Hooker, 1988)** *The construction of Proposition 3.12 generalizes conventional resolution.*

**Proof.** We have already discussed the case where the sets of $l_i$ and $l_i'$ are disjoint. If there is a literal $l_i$ in $c$ that is the negation of a literal in $c'$, then we will have $l_i + \neg l_i$ in (25), which we can simplify to 1 to make the resolved constraint trivial; resolution produces the same result. If there is a literal $l_i$ in $c$ that also appears in $c'$, the coefficient of that literal in the resolvent (25) will be 2 but can be reduced to 1 by virtue of the lemma.  ▫

Cardinality constraints are a bit more interesting. Suppose that we are resolving the two clauses

$$
\begin{array}{ccccccccc}
a & + & b & + & c & & & \geq & 2 \\
a & + & & & \overline{c} & + & d & \geq & 1
\end{array}
$$

which we add to get

$$2a + b + d \geq 2 \tag{26}$$

In other words, either $a$ is true or $b$ and $d$ both are. The problem is that this is not a cardinality constraint, and cannot be rewritten as one.

One possibility is to rewrite (26) as a pair of cardinality constraints

$$a + b \ \geq \ 1 \tag{27}$$
$$a + d \ \geq \ 1 \tag{28}$$

If, however, we want the result of "resolving" a pair of constraints to be a single axiom, we must either select one of the above axioms or extend our language further.

### 3.3 Learning and Relevance Bounds

The idea of relevance also has a natural generalization to the pseudo-Boolean setting. Recall the basic definition from Section 2.2:

**Definition 3.15** *Let $c$ be a clause and $P$ a partial assignment. Then $c$ is $i$-irrelevant if the number of literals in $c$ that are either unvalued or true under $P$ is at least $i + 1$.*

**Proposition 3.16** *Given a partial assignment $P$ and a Boolean clause $c$, $c$ is $i$-irrelevant if and only if $\mathtt{poss}(c, P) \geq i$.*

**Proof.** In the Boolean case, the number of literals in $c$ that are either unvalued or true is $\mathtt{poss}(c, P) + 1$ since the right hand side of the constraint is always 1. So the irrelevance condition is

$$\mathtt{poss}(c, P) + 1 \geq i + 1$$

and the result follows.  ▫

In the pseudo-Boolean case, additional learning techniques are also possible. Before we present these ideas in detail, however, let us point out that *some* sort of inferential extension is needed if we are to overcome the shortcomings of DPLL as revealed by the pigeonhole and other problems. After all, recall Proposition 3.14: pseudo-Boolean inference generalizes Boolean resolution. So if we begin with a Boolean axiomatization (as we did in the pigeonhole problem), any derivation using our techniques will be reproducible using conventional resolution-based methods, and will therefore be of exponential length. (A majority of the inference steps in the various proofs of Proposition 3.2 are not resolution steps in that no literal cancellations occur.)

**Strengthening** The specific method that we will discuss is from operations research and is used to preprocess mixed integer programming problems (Guignard & Spielberg, 1981; Savelsbergh, 1994).

Suppose that after setting $l_0$ to true and applying some form of propagation to our constraint set, we discover that under this assumption a constraint $c$ given by $\sum w_i l_i \geq r$ becomes oversatisfied by an amount $s$ in that the sum of the left hand side is greater (by $s$) than the amount required by the right hand side of the inequality; in the terms of Definition 3.5, $\texttt{curr}(c) = s$. The oversatisfied constraint $c$ can now be replaced by the following:

$$s\bar{l}_0 + \sum w_i l_i \geq r + s \qquad (29)$$

If $l_0$ is true, we know that $\sum w_i l_i \geq r + s$, so (29) holds. If $l_0$ is false, then $s\bar{l}_0 = s$ and we still must satisfy the original constraint $\sum w_i l_i \geq r$, so (29) still holds. The new constraint implies the original one, so no information is lost in the replacement.

As we have remarked, the OR community uses this technique during preprocessing. A literal is fixed, propagation is applied, and any oversatisfied constraint is strengthened. Consider the following set of clauses:

$$a + b \geq 1$$
$$a + c \geq 1$$
$$b + c \geq 1$$

If we set $a$ to false, we must then value both $b$ and $c$ true in order to satisfy the first two constraints. The third constraint is now oversatisfied and can thus be replaced by

$$a + b + c \geq 2$$

The power of this method is that it allows us to build more complex axioms from a set of simple ones. The strengthened constraint will often subsume some or all of the constraints involved in generating it. In this case, the new constraint subsumes all three of the generating constraints.

**Proposition 3.17** *Let $c$ be a constraint and $P$ a partial assignment. Then if we can conclude that $\texttt{curr}(c) \geq s$ for any solution to our overall problem that extends $P$, we can replace $c$ with*

$$s \sum_P \bar{l}_i + \sum w_i l_i \geq r + s \qquad (30)$$

*where the first summation is over literals $l_i \in P$.*

**Proof.** For any truth assignment that extends $P$, (30) follows from the fact that $\texttt{curr}(c) \geq s$. For any truth assignment $P'$ that does not extend $P$, there is some $l_j \in P$ that is false in $P'$, and so

$$s \sum_P \bar{l}_i \geq s$$

Combining this with the original constraint $c$ once again produces (30). $\quad\Box$

219

| Instance | zCHAFF | | PBCHAFF | |
|---|---|---|---|---|
| | sec | nodes | sec | nodes |
| 2pipe | 0 | 8994 | 0 | 8948 |
| 2pipe-1-ooo | 1 | 10725 | 1 | 9534 |
| 2pipe-2-ooo | 0 | 6690 | 0 | 6706 |
| 3pipe | 7 | 48433 | 12 | 57218 |
| 3pipe-1-ooo | 6 | 33570 | 9 | 36589 |
| 3pipe-2-ooo | 9 | 41251 | 16 | 45003 |
| 3pipe-3-ooo | 11 | 46504 | 19 | 57370 |
| 4pipe | 244 | 411107 | 263 | 382750 |

Table 1: Run time (seconds) and nodes expanded

**Learning and inference** Before we present some experimental results related to the effectiveness of pseudo-Boolean inference, we should point out one additional problem that can arise in this setting. It is possible that for some branch variable $v$, the result of resolving the reasons for $v$ and $\neg v$ is a new nogood that is not falsified by the partial assignment above $v$ in the search space.

As an example (Dixon & Ginsberg, 2002), suppose that we have a partial assignment $\{\neg a, \neg b, c, d\}$ and constraints

$$2e + a + c \quad \geq \quad 2 \tag{31}$$

$$2\overline{e} + b + d \quad \geq \quad 2 \tag{32}$$

Now we can unit propagate to conclude $e$ by virtue of (31) and $\neg e$ by virtue of (32); it isn't hard to conclude that the conflict set is $a \vee b$ in that either $a$ or $b$ must be true if (31) and (32) are to be simultaneously satisfiable. But if we simply add (31) and (32) and simplify, we get

$$a + b + c + d \geq 2$$

which still allows $a$ and $b$ to both be false. This difficulty can be addressed by deriving a cardinality constraint that is guaranteed to be falsified by the current partial solution being investigated (Dixon & Ginsberg, 2002); Chai and Kuehlmann (2003) have developed a still stronger method.

**Experimental results** Many of the ideas that we have described have been implemented in the PBCHAFF satisfiability solver. In an earlier paper (Dixon & Ginsberg, 2002), we compared results obtained using PRS, a pseudo-Boolean version of RELSAT, and those obtained using RELSAT (Bayardo & Miranker, 1996). PBCHAFF is an updated version of PRS that is modeled closely on zCHAFF (Moskewicz et al., 2001). It implements watched literals for cardinality constraints and applies the strengthening idea. Here we compare PBCHAFF's performance to its Boolean counterpart zCHAFF.

Results on some (unsatisfiable) problem instances from the Velev suite discussed at the beginning of Section 2.1 are shown in Table 1. As can be seen, performance is comparable; PBCHAFF pays a small (although noticeable) cost for its extended expressivity. The experiments were run on a 1.5 GHz AMD Athlon processor, and both solvers used the same values for the various tuning parameters available (relevance and length bounds, etc.).

| Instance | zCHAFF | | Preprocess | PBCHAFF | |
|---|---|---|---|---|---|
| | sec | nodes | sec | sec | nodes |
| hole8.cnf | 0 | 3544 | 0 | 0 | 11 |
| hole9.cnf | 1 | 8144 | 0 | 0 | 12 |
| hole10.cnf | 17 | 27399 | 0 | 0 | 17 |
| hole11.cnf | 339 | 126962 | 0 | 0 | 15 |
| hole12.cnf | | | 0 | 0 | 20 |
| hole20.cnf | | | 0 | 0 | 34 |
| hole30.cnf | | | 4 | 0 | 52 |
| hole40.cnf | | | 25 | 0 | 75 |
| hole50.cnf | | | 95 | 0 | 95 |

Table 2: Run time (seconds) and nodes expanded

Results for the pigeonhole problem appear in Table 2. In this case, PBCHAFF was permitted to preprocess the problem using strengthening as described earlier in this section. ZCHAFF was unable to solve the problem for twelve or more pigeons with a 1000-second timeout using a 1.5 GHz Athlon processor. Not surprisingly, PBCHAFF with preprocessing dramatically outperformed zCHAFF on these instances.[13]

## 3.4 Proof Complexity

We have already shown in Proposition 3.2 that pseudo-Boolean or cardinality-based axiomatizations can produce polynomially sized proofs of the pigeonhole problem. It is also known that these methods do not lead to polynomially sized proofs of the clique coloring problem (Bonet et al., 1997; Kraj'icek, 1997; Pudlak, 1997). The situation with regard to parity constraints is a bit more interesting.

Let us first point out that it is possible to capture parity constraints, or modularity constraints generally in a pseudo-Boolean setting:

**Definition 3.18** *A* modularity constraint *is a constraint c of the form*

$$\sum_i w_i l_i \equiv n (\mathrm{mod}\ m) \tag{33}$$

*for positive integers $w_i$, $n$ and $m$.*

In the remainder of this section, we show that modularity constraints can be easily encoded using pseudo-Boolean axioms, and also that constraint sets consisting entirely of mod 2 constraints are easily solved either directly or using the above encoding, although it is not clear how to recover the pseudo-Boolean encodings from the Boolean versions.

---

13. Without preprocessing, the two systems perform comparably on this class of problems. As we have stressed, representational extensions are of little use without matching modifications to inference methods.

**Modularity constraints and pseudo-Boolean encodings**  To encode a modularity constraint in this way, we first note that we can easily capture an equality axiom of the form

$$\sum_i w_i l_i = k \tag{34}$$

in a pseudo-Boolean setting, simply by rewriting (34) as the pair of constraints

$$\sum_i w_i l_i \ \geq \ k$$

$$\sum_i w_i \bar{l}_i \ \geq \ \sum_i w_i - k$$

In what follows, we will therefore feel free to write axioms of the form (34).

We now denote by $\lfloor x \rfloor$ the floor of $x$, which is to say the smallest integer not greater than $x$, and have:

**Proposition 3.19** *Suppose that we have a modularity constraint of the form (33). We set $w = \sum_i w_i$ and introduce new variables $s_i$ for $i = 1, \ldots, \lfloor \frac{w}{m} \rfloor$. Then (33) is equivalent to*

$$\sum_i w_i l_i + \sum_i m s_i = m \left\lfloor \frac{w}{m} \right\rfloor + n \tag{35}$$

**Proof.** Reducing both sides of (35) mod m shows that (35) clearly implies (33). For the converse, note if (33) is satisfied, there is some integer $s$ such that $\sum_i w_i l_i = sm + n$. Further, since $\sum_i w_i l_i \leq \sum_i w_i = w$, it follows that $sm + n \leq w$, so that $s \leq \frac{w-n}{m} \leq \frac{w}{m}$ and thus $s \leq \lfloor \frac{w}{m} \rfloor$. We can therefore satisfy (35) by valuing exactly that many of the $s_i$ to be true.  □

Understand that the introduction of new variables here is not part of any intended inference procedure; it is simply the fashion in which the modularity constraints can be captured within a pseudo-Boolean setting.

In the case where all of the constraints are parity constraints, we have:

**Proposition 3.20** *A set of mod 2 constraints can be solved in polynomial time.*

**Proof.** An individual constraint (recall that $\oplus$ corresponds to exclusive or, or addition mod 2)

$$l \oplus \oplus_i l_i = n$$

can be viewed simply as defining

$$l = n \oplus \oplus_i l_i$$

and this definition can be inserted to remove $l$ from the remaining constraints. Continuing in this way, we either define all of the variables (and can then return a solution) or derive $1 = 0$ and can return failure.  □

This result, which can be thought of as little more than an application of Gaussian elimination, is also an instance of a far more general result of Schaefer's (1978).

**Proposition 3.21** *A set of mod 2 constraints can be solved in polynomial time using the pseudo-Boolean axiomatization given by (35).*

**Proof.** The technique is unchanged. When we combine

$$l + \sum_i l_i + 2 \sum_i s_i = n$$

and

$$l + \sum_i l_i' + 2 \sum_i s_i' = n'$$

we get

$$\sum_i l_i + \sum_i l_i' + 2(\sum_i s_i + \sum_i s_i' + l) = n + n'$$

and can now treat $l$ as one of the auxiliary $s$ variables. Eventually, we will get

$$2 \sum_i s_i = n$$

for a large (but polynomially sized) set $S$ of auxiliary variables and some $n$ that is either even or odd. If $n$ is even, we can value the variables and return a solution; if $n$ is odd and there are $k$ auxiliary variables, we have

$$\sum_i s_i = \frac{n}{2}$$

so

$$\sum_i s_i \geq \frac{n+1}{2} \tag{36}$$

since each $s_i$ is integral. But we also have

$$2 \sum_i \overline{s}_i \geq 2k - n$$

so that

$$\sum_i \overline{s}_i \geq k - \frac{n-1}{2} \tag{37}$$

Adding (36) and (37) produces $k \geq k + 1$, a contradiction. □

Let us point out, however, that if a mod 2 constraint is encoded in a normal Boolean way, so that $x \oplus y \oplus z = 1$ becomes

$$x \vee y \vee z \tag{38}$$
$$x \vee \neg y \vee \neg z$$
$$\neg x \vee y \vee \neg z$$
$$\neg x \vee \neg y \vee z \tag{39}$$

it is not obvious how the pseudo-Boolean analog can be reconstructed. Here is the problem we mentioned at the beginning of this section: it is not enough to simply extend the representation; we need to extend the inference methods as well. In fact, even the question of whether families of mod 2 constraints can be solved in polynomial time by pseudo-Boolean methods without the introduction of auxiliary variables as in (35) is open. Other authors have also considered the problem of reasoning with these constraints directly (Li, 2000).

**Pseudo-Boolean constraints and extended resolution**  Finally, let us clarify a point that we made earlier. Given that there is an encoding of $a \lor (b \land c)$ as a single pseudo-Boolean clause, how can it be that pseudo-Boolean inference is properly below extended resolution in the $p$-simulation hierarchy?

The answer is as follows. While the fact that $a \lor (b \land c)$ is logically equivalent to $2a + b + c \geq 2$ allows us to remove one of the variables introduced by extended resolution, we cannot combine this encoding with others to remove subsequent variables. As a specific example, suppose that we learn both

$$a \lor (b \land c)$$

and

$$d \lor (b \land c)$$

and wish to conclude from this that

$$(a \land d) \lor (b \land c) \tag{40}$$

There is no single pseudo-Boolean axiom that is equivalent to (40).

## 3.5 Summary

| | rep. eff. | p-simulation hierarchy | inference | unit propagation | learning |
|---|---|---|---|---|---|
| **SAT** | 1 | EEE | resolution | watched literals | relevance |
| **cardinality** | exp | P?E | not unique | watched literals | relevance |
| **PB** | exp | P?E | uniquely defined | watched literals | + strengthening |
| **symmetry** | | | | | |
| **QPROP** | | | | | |

As before, a few notes are in order:

- While both cardinality and pseudo-Boolean representations can be exponentially more efficient than their Boolean counterpart, it is not clear how often compressions of this magnitude will occur in practice.

- The entries in the $p$-simulation column indicate that the pigeonhole problem is easy, clique coloring remains hard, and the complexity of parity problems is unknown if no new variables are introduced.

- The cardinality entry for "inference" is intended to reflect the fact that the natural resolvent of two cardinality constraints need not be one.

- Pseudo-Boolean systems can use existing learning techniques, augmented with the strengthening idea.

## 4. Symmetry

Given that the pigeonhole problem and clique-coloring problems involve a great deal of symmetry in their arguments, a variety of authors have suggested extending Boolean representation or inference in a way that allows this symmetry to be exploited directly. We will discuss the variety of approaches that have been proposed by separating them based on whether or not a modification to the basic resolution inference rule is suggested. In any event, we make the following definition:

**Definition 4.1** *Let $T$ be a collection of axioms. By a* symmetry *of $T$ we will mean any permutation $\rho$ of the variables in $T$ that leaves $T$ itself unchanged.*

As an example, if $T$ consists of the single axiom $x \vee y$, then $T$ is clearly symmetric under the exchange of $x$ and $y$. If $T$ contains the two axioms

$$a \vee x$$

and

$$a \vee y$$

then $T$ is once again symmetric under the exchange of $x$ and $y$.

**Exploiting symmetry without changing inference** One way to exploit symmetry is to modify the set of axioms in a way that captures the power of the symmetry. In the pigeonhole problem, for example, we can argue that since there is a symmetry under the exchange of pigeons or of holes, we can assume "without loss of generality" that pigeon 1 is in hole 1, and then by virtue of a residual symmetry that pigeon 2 is in hole 2, and so on.

The basic idea is to add so-called *symmetry-breaking axioms* to our original theory, axioms that break the existing symmetry without affecting the overall satisfiability of the theory itself. This idea was introduced by Crawford et al. (1996).

While attractive in theory, there are at least two fundamental difficulties with the symmetry-breaking approach:

1. Luks and Roy (2002) have shown that breaking all of the symmetries in any particular problem may require the introduction of a set of symmetry-breaking axioms of exponential size. This problem can be sidestepped by breaking only "most" of the symmetries, although little is known about how the set of broken symmetries is to be selected.

2. Far more serious, the technique can only be applied if the symmetry in question is global. This is because the basic argument that satisfiability is unaffected by the introduction of the new axioms requires that there be no additional axioms to consider.

In theoretical problems, global symmetries exist. But in practice, even the addition of asymmetric axioms that constrain the problem further (e.g., you can't put pigeon 4 in hole 7) will break the required global symmetry and render this method inapplicable. More problematic still is the possibility of the symmetries being "obscured" by replacing the single axiom

$$\neg p_{11} \vee \neg p_{21} \tag{41}$$

with the equivalent pair

$$a \vee \neg p_{11} \vee \neg p_{21}$$

and

$$\neg a \vee \neg p_{11} \vee \neg p_{21}$$

from which (41) can obviously be recovered using resolution. Once again, the symmetry in the original problem has vanished and the method cannot be applied.

These arguments could perhaps have been anticipated by consideration of our usual table; since the inference mechanism itself is not modified (and it is possible to break global symmetries), none of the entries has changed. Let us turn, then, to other techniques that modify inference itself.

**Exploiting symmetry by changing inference**   Rather than modifying the set of clauses in the problem, it is also possible to modify the notion of inference, so that once a particular nogood has been derived, symmetric equivalents can be derived in a single step. The basic idea is due to Krishnamurthy (1985) and is as follows:

**Lemma 4.2** *Suppose that* $T \models q$ *for some theory* $T$ *and nogood* $q$. *If* $\rho$ *is a symmetry of* $T$, *then* $T \models \rho(q)$. □

It is not hard to see that this technique allows the pigeonhole problem to be solved in polynomial time, since symmetric versions of specific conclusions (e.g., pigeon 1 is not in hole 1) can be derived without repeating the analysis that led to the original. The dependence on global symmetries remains, but can be addressed by the following modification:

**Proposition 4.3** *Let* $T$ *be a theory, and suppose that* $T' \models q$ *for some* $T' \subseteq T$ *and nogood* $q$. *If* $\rho$ *is a symmetry of* $T'$, *then* $T \models \rho(q)$. □

Instead of needing to find a symmetry of the theory $T$ in its entirety, it suffices to find a "local" symmetry of the subset of $T$ that was actually used in the proof of $q$.

This idea, which has been generalized somewhat by Szeider (2003), allows us to avoid the fact that the introduction of additional axioms can break a global symmetry. The problem of symmetries that have been obscured as in (41) remains, however, and is accompanied by a new one, the need to identify local symmetries at each inference step (Brown et al., 1988).

While it is straightforward to identify the support of any new nogood $q$ in terms of a subtheory $T'$ of the original theory $T$, finding the symmetries of any particular theory is equivalent to the graph isomorphism problem (Crawford, 1992). The precise complexity of graph isomorphism is unknown, but it is felt likely to be properly between $P$ and $NP$ (Babai, 1995). Our basic table becomes:

| | rep. eff. | p-simulation hierarchy | inference | unit propagation | learning |
|---|---|---|---|---|---|
| **SAT** | 1 | EEE | resolution | watched literals | relevance |
| **cardinality** | exp | P?E | not unique | watched literals | relevance |
| **PB** | exp | P?E | unique | watched literals | + strengthening |
| **symmetry** | 1 | EEE* | not in P | same as SAT | same as SAT |
| **QPROP** | | | | | |

- It is not clear how the representational efficiency of this system is to be described, since a single concluded nogood can serve as a standin for its image under the symmetries of the proof that produced it.

- While specific instances of the pigeonhole problem and clique coloring problems can be addressed using symmetries, even trivial modifications of these problems render the techniques inapplicable. Hence the appearance of the asterisk in the above table: "Textbook" problem instances may admit polynomially sized proofs, but most instances require proofs of exponential length. Parity problems do not seem to be amenable to these techniques at all.

- As we have remarked, inference using Krishnamurthy's or related ideas appears to require multiple solutions of the graph isomorphism problem, and is therefore unlikely to remain in $P$.

We know of no implemented system based on the ideas discussed in this section.

## 5. Quantification and QPROP

We conclude our survey with an examination of ideas that have been used in trying to extend the Boolean work to cope with theories that are most naturally thought of using quantification of some sort. Indeed, as Boolean satisfiability engines are applied to ever larger problems, many of the theories in question are produced in large part by constructing the set of ground instances of quantified axioms such as

$$\forall xyz.[a(x, y) \land b(y, z) \to c(x, z)] \tag{42}$$

If $d$ is the size of the domain from which $x$, $y$ and $z$ are taken, this single axiom has $d^3$ ground instances. Researchers have dealt with this difficulty by buying machines with more memory or by finding clever axiomatizations for which ground theories remain manageably sized (Kautz & Selman, 1998). In general, however, memory and cleverness are both scarce resources and a more natural solution needs to be found.

We will call a clause such as (42) *quantified*, and assume throughout this section that the quantification is universal as opposed to existential, and that the domains of quantification are finite.[14]

As we remarked at the beginning of this section, quantified clauses are common in encodings of realistic problems, and these problems have in general been solved by converting quantified clauses to standard propositional formulae. The quantifiers are expanded first (possible because the domains of quantification are finite), and the resulting set of predicates is then "linearized" by relabeling all of the atoms so that, for example, $a(2, 3)$ might become $v_{24}$. The number of ground clauses produced is exponential in the number of variables in the quantified clause.

---

14. There appears to be no effective alternative but to treat existentially quantified clauses as simple disjunctions, as in (9).

## 5.1 Unit Propagation

Our primary goal here is to work with the quantified formulation directly, as opposed to its much larger ground translation. Unfortunately, there are significant constant-factor costs incurred in doing so, since each inference step will need to deal with issues involving the bindings of the variables in question. Simply finding the value assigned to $a(2, 3)$ might well take several times longer than finding the value assigned to the equivalent $v_{24}$. Finding all occurrences of a given literal can be achieved in the ground case by simple indexing schemes, whereas in the quantified case this is likely to require a unification step. While unification can be performed in time linear in the length of the terms being unified, it is obviously not as efficient as a simple equality check. Such routine but essential operations can be expected to significantly slow the cost of every inference undertaken by the system.

Our fundamental point here is that while there are costs associated with using quantified axioms, there are significant savings as well. These savings are a consequence of the fact that the basic unit propagation procedure uses an amount of time that scales roughly linearly with the size of the theory; use of quantified axioms can reduce the size of the theory so substantially that the constant-factor costs can be overcome.

We will make this argument in two phases. In Section 5.1.1, we generalize a specific computational subtask that is shared by unit propagation and other satisfiability procedures such as WSAT. We will show this generalization to be NP-complete in a formal sense, and we call it *subsearch* for that reason. The specific procedures for unit propagation and as needed by WSAT encounter this NP-complete subproblem at each inference step, and we show that while subsearch is generally not a problem for randomly generated theories, the subsearch cost can be expected to dominate the running time on more realistic instances.

In Section 5.1.2, we discuss other consequences of the fact that subsearch is NP-complete. Search techniques can be used to speed the solution of NP-complete problems, and subsearch is no exception. We show that quantified axiomatizations support the application of simple search techniques to the subsearch problem, and argue that realistic examples are likely to lead to subsearch problems of only polynomial difficulty although existing unit propagation implementations solve them exponentially.

### 5.1.1 SUBSEARCH

Each iteration of DPLL (or WSAT) involves a search through the original theory for clauses that satisfy some numeric property. The specific examples that we have already seen of this are the following:

1. In Procedure 2.2 (DPLL) (and similarly in WSAT), we need to determine if $P$ is a solution to the problem at hand. This involves searching for an unsatisfied clause.

2. In Procedure 2.3 (unit propagation), we need to find unsatisfied clauses that contain at most one unvalued literal.

In addition, WSAT needs to compute the number of clauses that will become unsatisfied when a particular variable is flipped.

All of these tasks can be rewritten using the following:

**Definition 5.1** *Suppose $C$ is a set of quantified clauses, and $P$ is a partial assignment of values to the atoms in those clauses. We will denote by $S_u^s(C, P)$ the set of ground instances of $C$ that have $u$ literals unvalued by $P$ and $s$ literals satisfied by the assignments in $P$.*[15]

*We will say that the* checking problem *is that of determining whether $S_u^s(C, P) \neq \varnothing$. By a* subsearch problem, *we will mean an instance of the checking problem, or the problem of either enumerating $S_u^s(C, P)$ or determining its size.*

**Proposition 5.2** *For fixed $u$ and $s$, the checking problem is NP-complete.*

**Proof.** Checking is in NP, since a witness that $S_u^s(C, P) \neq \varnothing$ need simply give suitable bindings for the variables in each clause of $C$.

To see NP-hardness, we assume $u = s = 0$; other cases are not significantly different. We reduce from a binary constraint satisfaction problem (CSP), producing a single clause $C$ and set of bindings $P$ such that $S_0^0(C, P) \neq \varnothing$ if and only if the original binary CSP was satisfiable. The basic idea is that each variable in the constraint problem will become a quantified variable in $C$.

Suppose that we have a binary CSP $\Sigma$ with variables $v_1, \ldots, v_n$ and with $m$ binary constraints of the form $(v_{i1}, v_{i2}) \in c_i$, where $(v_{i1}, v_{i2})$ is the pair of variables constrained by $c_i$. For each such constraint, we introduce a corresponding binary relation $r_i(v_{i1}, v_{i2})$, and take $C$ to be the single quantified clause $\forall v_1, \ldots, v_n. \vee_i r_i(v_{i1}, v_{i2})$. For the assignment $P$, we set $r_i(v_{i1}, v_{i2})$ to false for all $(v_{i1}, v_{i2}) \in c_i$, and to true otherwise.

Now note that since $P$ values every instance of every $r_i$, $S_0^0(C, P)$ will be nonempty if and only if there is a set of values for $v_i$ such that every literal in $\vee_i r_i(v_{i1}, v_{i2})$ is false. Since a literal $r_i(v_{i1}, v_{i2})$ is false just in the case the original constraint $c_i$ is satisfied, it follows that $S_0^0(C, P) \neq \varnothing$ if and only if the original CSP $\Sigma$ was satisfiable. $\square$

Before moving on, let us place this result in context. First, and most important, note that the fact that the checking problem is NP-complete does not imply that QPROP is an unwieldy representation; the subsearch problem does indeed appear to be exponential in the size of the QPROP axioms, but there are exponentially fewer of them than in the ground case. So, as for similar results elsewhere (Galperin & Wigderson, 1983; Papadimitriou, 1994), there is no net effect on complexity.

Second, the result embodied in Proposition 5.2 appears to be a general phenomenon in that propagation is more difficult for more compact representations. Our earlier discussion of cardinality and pseudo-Boolean axioms, for which the complexity of unit propagation was unchanged from the Boolean case, appears to be much more the exception than the rule. As we have already remarked, if we extend the pseudo-Boolean representation only slightly, so that in addition to axioms of the form

$$\sum_i w_i l_i \geq k \tag{43}$$

as in Definition 3.4 we allow axioms such as

$$\sum_i w_i l_i = k$$

---

15. In interpreting the expression $S_u^s(C, P)$, the set $C$ of clauses and partial assignment $P$ should generally be clear from context. The superscript refers to the number of satisfied literals because satisfied literals are "super good" and the subscript refers to the unvalued literals because unvalued literals aren't so good.

(replacing the inequality in (43) with an equality), determining whether a single axiom is satisfiable becomes weakly NP-complete. Symmetry, the other example we have examined, involves no effective change in the representational power of a single axiom.

Here is a recasting of unit propagation in terms of Definition 5.1:

**Procedure 5.3 (Unit propagation)**  *To compute* Unit-Propagate($P$):

1   **while** $S_0^0(C, P) = \emptyset$ **and** $S_1^0(C, P) \neq \emptyset$
2        **do** select $c \in S_1^0(C, P)$
3             $v \leftarrow$ the variable in $c$ unassigned by $P$
4             $P \leftarrow P \cup \{v = V : V \text{ is selected so that } c \text{ is satisfied}\}$
5   **return** $P$

It is important to recognize that this recasting is not changing the procedure in any significant way; it is simply making explicit the subsearch tasks that were previously described only implicitly. The procedure itself is unchanged, and other procedural details such as variable and value choice heuristics are irrelevant to the general point that unit propagation depends on solving a subsearch instance at every step. Wsat is similar.

### 5.1.2 Subsearch and quantification

As we discussed in Section 2.1, efficient implementations of sat solvers go to great lengths to minimize the amount of time spent solving subsearch problems. While the watched literal idea is the most efficient mechanism known here, we will discuss the problem in terms of a simpler scheme that maintains and updates `poss` and `curr` counts for each clause. As discussed earlier, this scheme is about half as fast as the watched literal approach, and the general arguments that we will make can be expected to lead to more than constant-factor improvements.[16]

For notational convenience in what follows, suppose that $C$ is a quantified theory and that $l$ is a ground literal. By $C_l$ we will mean that subset of the clauses in $C$ that include terms of which $l$ is an instance. If $C$ contains quantified clauses, then $C_l$ will as well; the clauses in $C_l$ can be found by matching the literal $l$ against the clauses in $C$.

As discussed in Section 2.1, it is possible to compute $S_u^s(C, P)$ once during an initialization phase, and then update it incrementally. In terms of Definition 5.1, the update rule might be one such as

$$S_0^0(C, P') = S_0^0(C, P) \cup S_1^0(C_{\neg l}, P)$$

if the literal $l$ is changed from unvalued to true. $P'$ here is the partial assignment after the update; $P$ is the assignment before. To compute the number of fully assigned but unsatisfied clauses after the update, we start with the number before, and add newly unsatisfied clauses (unsatisfied clauses previously containing the single unvalued literal $\neg l$).

As we argued previously, reorganizing the computation in this way leads to substantial speedups because the subsearch problem being solved is no longer NP-complete in the size of $C$, but only in the size of $C_l$ or $C_{\neg l}$. These incremental techniques are essential to the performance of modern search implementations because the runtime of these implementations is dominated by the time spent in propagation (i.e., subsearch).

---

16. We know of no effective way to lift the watched literal idea to the qprop setting. But as we will see when we discuss the zap implementation (Dixon et al., 2003a), a still broader generalization allows watched literals to return in an elegant and far more general way.

Given that the subsearch computation time is potentially exponential in the size of the subtheory $C_l$ when the literal $l$ is valued or unvalued, let us now consider the questions of how much of a concern this is in practice, and of what (if anything) can be done about it. After all, one of the primary lessons of recent satisfiability research is that problems that are NP-hard in theory tend strongly not to be exponentially difficult in practice.

Let us begin by noting that subsearch is *not* likely to be much of an issue for the randomly generated satisfiability problems that were the focus of research in the 1990's and drove the development of algorithms such as WSAT. The reason for this is that if $n$ is the number of clauses in a theory $C$ and $v$ is the number of variables in $C$, then random problems are difficult only for fairly narrow ranges of values of the ratio $n/v$ (Coarfa, Demopoulos, San Miguel Aguirre, Subramanian, & Vardi, 2000). For 3-SAT (where every clause in $C$ contains exactly three literals), difficult random problems appear at $n/v \approx 4.2$ (Kirkpatrick & Selman, 1994). For such a problem, the number of clauses in which a particular literal $l$ appears will be small (on average $3 \times 4.2/2 = 6.3$ for random 3-SAT). Thus the size of the relevant subtheory $C_l$ or $C_{\neg l}$ will also be small, and while subsearch cost still tends to dominate the running time of the algorithms in question, there is little to be gained by applying sophisticated techniques to reduce the time needed to examine a relative handful of clauses.

For realistic problems, the situation is dramatically different. Here is an axiom from a logistics domain encoded in SATPLAN style (Kautz & Selman, 1992):

$$\texttt{at}(o, l, t) \wedge \texttt{duration}(l, l', dt) \wedge$$
$$\texttt{between}(t, t', t + dt) \rightarrow \neg\texttt{at}(o, l', t') \tag{44}$$

This axiom says that if an object $o$ is at location $l$ at time $t$ and it takes time $dt$ to fly from $l$ to $l'$, and $t'$ is between $t$ and $t + dt$, then $o$ cannot be at $l'$ at $t'$.

A given ground atom of the form $\texttt{at}(o, l, t)$ will appear in $|t|^2 |l|$ clauses of the above form, where $|t|$ is the number of time points or increments and $|l|$ is the number of locations. Even if there are only 100 of each, the $10^6$ axioms created seem likely to make computing $S_u^s(C_l, P)$ impractical.

Let us examine this computation in a bit more detail. Suppose that we do indeed have a variable $a = \texttt{at}(O, L, T)$ for fixed $O$, $L$ and $T$, and that we are interested in counting the number of unit propagations that will be possible if we set $a$ to true. In other words, we want to know how many instances of (44) will be unsatisfied and have a single unvalued literal after we do so.

Existing implementations, faced with this problem (or an analogous one if WSAT or another approach is used), will now consider axioms of the form (44) for $o$, $l$ and $t$ bound and as $l'$, $t'$ and $dt$ are allowed to vary. They examine every axiom of this form and simply count the number of possible unit propagations.

The watched literal idea in isolation cannot help with this problem. If, for example, we watch only the $\texttt{duration}$ and $\texttt{between}$ predicates in (44), we reduce by half the probability that we need to solve a subsearch problem when a particular variable is valued, but in those cases where the problem is encountered, it is as fierce as ever.

The existing approach to solving subsearch problems is taken because existing systems use not quantified clauses such as (44), but the set of ground instances of those clauses. Computing $S_u^s(C, P)$ for ground $C$ involves simply checking each axiom individually; indeed, once the axiom has been replaced by its set of ground instances, no other approach seems possible.

Set against the context of a quantified axiom, however, this seems inappropriate. Computing $S_u^s(C, P)$ for a quantified $C$ by reducing $C$ to a set of ground clauses and then examining each is equivalent to solving the original NP-complete problem by generate and test – and if there is one thing that we can state with confidence about NP-complete problems, it is that generate and test is not in general an effective way to solve them.

Returning to our example with $\mathtt{at}(O, L, T)$ true, we are looking for variable bindings for $l'$, $dt$ and $t'$ such that, amongst $\neg\mathtt{duration}(L, l', dt)$, $\neg\mathtt{between}(T, t', T + dt)$ and $\neg\mathtt{at}(O, l', t')$, precisely two of these literals are false and the third is unvalued. Proposition 5.2 suggests that subsearch will be exponentially hard (with respect to the number of quantifiers) in the worst case, but what is it likely to be like in practice?

In practice, things are going to be much better. Suppose that for some possible destination $l'$, we know that $\mathtt{duration}(L, l', dt)$ is false for all $dt$ except some specific value $D$. We can immediately ignore all bindings for $dt$ except for $dt = D$, reducing the size of the subsearch space by a factor of $|t|$. If $D$ depended on previous choices in the search (aircraft loads, etc.), it would be impossible to perform this analysis in advance and thereby remove the unnecessary bindings in the ground theory.

Pushing this example somewhat further, suppose that $D$ is so small that $T + D$ is the time point immediately after $T$. In other words, $\mathtt{between}(T, t', T + D)$ will always be false, so that $\neg\mathtt{between}(T, t', T + D)$ will always be true and no unit propagation will be possible for any value of $t'$ at all. We can "backtrack" away from the unfortunate choice of destination $l'$ in our (sub)search for variable bindings for which unit propagation is possible. Such backtracking is not supported by the generate-and-test subsearch philosophy used by existing implementations.

This sort of computational savings is likely to be possible in general. For naturally occurring theories, most of the variables involved are likely to be either unvalued (because we have not yet managed to determine their truth values) or false (by virtue of the closed-world assumption, Reiter, 1978, if nothing else). Domain constraints will typically be of the form $a_1 \wedge \cdots \wedge a_k \rightarrow l$, where the premises $a_i$ are variables and the conclusion $l$ is a literal of unknown sign. Unit propagation (or other likely instances of the subsearch problem) will thus involve finding a situation where at most one of the $a_i$ is unvalued, and the rest are true. If we use efficient data structures to identify those instances of relational expressions that are true, it is not unreasonable to expect that most instances of the subsearch problem will be soluble in time polynomial in the length of the clauses involved, as opposed to exponential in that length.

## 5.2 Inference and Learning

As in Section 3, working with a modified representation allows certain inference techniques that are not applicable in the Boolean case.

As an example, suppose that we are resolving

$$\neg a(A, B) \vee \neg b(B, C) \vee c(C)$$

with

$$\neg c(C) \vee d(C, D)$$

to conclude

$$\neg a(A, B) \vee \neg b(B, C) \vee d(C, D) \tag{45}$$

where the capital letters indicate ground elements of the domain as before and the resolvents are actually ground instances of

$$\neg a(x, y) \vee \neg b(y, C) \vee c(C) \tag{46}$$

and

$$\neg c(z) \vee d(z, w) \tag{47}$$

It is obviously possible to resolve (46) and (47) directly to obtain

$$\neg a(x, y) \vee \neg b(y, C) \vee d(C, w) \tag{48}$$

which is more general than (45). For a procedure that learns new nogoods and uses them to prune the resulting search, the impact of learning the more general (48) can be substantial and can easily outweigh the cost of the unification step required to conclude that $c(C)$ and $\neg c(z)$ resolve if $z = C$. We have also discussed this elsewhere (Parkes, 1999).

There are two new difficulties that arise when we implement these ideas. The first is a consequence of the fact that resolution can be ambiguously defined for two quantified clauses. Consider resolving

$$a(A, x) \vee a(y, B) \tag{49}$$

with

$$\overline{a}(A, B) \vee b(A, B) \tag{50}$$

If we unify the first term in (50) with the first term in (49), we obtain $a(y, B) \vee b(A, B)$ as the resolvent; if we unify with the second term of (49), we obtain $a(A, x) \vee b(A, B)$.

In practice, however, this need not be a problem:

**Proposition 5.4** *Let $c_1$ and $c_2$ be two lifted clauses, and $g_1$ and $g_2$ ground instances that resolve to produce $g$. Then there is a unique natural resolvent of $c_1$ and $c_2$ of which $g$ is a ground instance.*

**Proof.** If there is more than one pair of resolving literals in $g_1$ and $g_2$ the result of the resolution will be vacuous, so we can assume that there is a single literal $l$ in $g_1$ with $\neg l$ in $g_2$. If $l$ is the $i$th literal in $g_1$ and $\neg l$ the $j$th literal in $g_2$, it follows that we can resolve the original $c_1$ and $c_2$ by unifying the $i$th literal in $c_1$ and the $j$th literal in $c_2$. It is clear that this resolution will be a generalization of $g$. □

What this suggests is that the reasons being associated with literal values be not the lifted nogoods that are retained as clauses, but ground instances thereof that were initially used to prune the search space and can subsequently be used to break ambiguities in learning.

The second difficulty is far more substantial. Suppose that we have the axiom

$$\neg a(x, y) \lor \neg a(y, z) \lor a(x, z)$$

or, in a more familiar form, the usual transitivity axiom

$$a(x, y) \land a(y, z) \to a(x, z)$$

This might be used in reasoning about a logistics problem, for example, if it gave conditions under which two cities were connected by roads.

Now suppose that we are trying to prove $a(A, B)$ for an $A$ and a $B$ that are "far apart" given the skeleton of the relation $a$ that we already know. It is possible that we use resolution to derive

$$a(A, x) \land a(x, B) \to a(A, B)$$

as we search for a proof involving a single intermediate location, and then

$$a(A, x) \land a(x, y) \land a(y, B) \to a(A, B)$$

as we search for a proof involving two such locations, and so on, eventually deriving the wonderfully concise

$$a(A, x_1) \land \cdots \land a(x_n, B) \to a(A, B) \tag{51}$$

for some suitably large $n$.

The problem is that if $d$ is the size of our domain, (51) will have $d^n$ ground instances and is in danger of overwhelming our unit propagation algorithm even in the presence of reasonably sophisticated subsearch techniques. Some technique needs to be adopted to ensure that this difficulty is sidestepped in practice. One way to do this is to learn not the fully general (51), but a partially bound instance that has fewer ground instances.

**Procedure 5.5** *To construct* `learn`$(c, g)$, *the nogood that will be learned after a clause $c$ has been produced in response to a backtrack, with $g$ the ground reason associated with $c$:*

1   **while** $c$ has a ground instance that is $i$-irrelevant
2       **do** $v \leftarrow$ a variable in $c$
3          bind $v$ to its value in $g$
4   **return** $c$

We may still learn a nogood with an exponential number of ground instances, but at least have some reason to believe that each of these instances will be useful in pruning subsequent search. Note that there is a subsearch component to Procedure 5.5, since we need to find ground instances of $c$ that are irrelevant. This cost is incurred only once when the clause is learned, however, and not at every unit propagation or other use.

It might seem more natural to learn the general (51), but to modify the subsearch algorithm used in unit propagation so that only a subset of the candidate clauses is considered. As above, the most natural approach would likely be to restrict the subsearch to clauses of a particular irrelevance or better. Unfortunately, this won't help, since irrelevant clauses cannot be unit. Restricting the subsearch to relevant clauses is no more useful in practice than requiring that any search algorithm expand only successful nodes.

Before moving on, let us note that a similar phenomenon occurs in the pseudo-Boolean case. Suppose we have a partial assignment $\{\neg b, c, \neg d, e\}$ and constraints

$$a + d + \overline{e} \geq 1 \tag{52}$$

$$\overline{a} + b + c \geq 2 \tag{53}$$

Unit propagation now causes the variable $a$ to be simultaneously true (by virtue of (52)) and false (because of (53)). Resolving these reasons together as in Proposition 3.12 gives us

$$b + c + d + \overline{e} \geq 2 \tag{54}$$

The conflict set here is easily seen to be $\{\neg b, \neg d, e\}$, and this is indeed prohibited by the derived constraint (54). But (54) eliminates some additional bad assignments as well, such as $\{\neg c, \neg d, e\}$. Just as in the lifted case, we have learned something about a portion of the search space that has yet to be examined.

## 5.3 Summary

| | rep. eff. | p-simulation hierarchy | inference | unit propagation | learning |
|---|---|---|---|---|---|
| **SAT** | 1 | EEE | resolution | watched literals | relevance |
| **cardinality** | exp | P?E | not unique | watched literals | relevance |
| **PB** | exp | P?E | unique | watched literals | + strengthening |
| **symmetry** | 1 | EEE* | not in P | same as SAT | same as SAT |
| **QPROP** | exp | ??? | in P using reasons | exp improvement | + first-order |

As usual, there are a few points to be made.

- There is an important difference in practice between the exponential savings in representation provided by QPROP and the savings provided by pseudo-Boolean or cardinality encodings. While the exponential savings in previous cases were mathematical possibilities that were of uncertain use in practice, the savings provided by QPROP can be expected to be achieved in any axiomatization that is constructed by grounding out a relative handful of universally quantified physical laws.

- It is not clear whether QPROP leads to polynomially sized solutions to the pigeonhole and clique coloring problems. It appears at first blush that it should, since quantification over pigeons or holes is the QPROP analog of the identification of the corresponding symmetry as in the previous section. We know of no detailed proof in the literature, however, and our attempts to construct one have been unsuccessful. Similar remarks apply to parity problems.

- Inference in QPROP requires the introduction of a (linear complexity) unification step, and is only uniquely defined if reasons are maintained for the choices made in the search.

- The exponential savings claimed for unit propagation are obviously an average case result, as opposed to a worst case one. They are a consequence of the fact that it is possible to use subsearch as part of unit propagation, as opposed to the "generate and test" mechanism used by ground methods.

- In addition to the usual idea of relevance-based learning, quantified methods can extend the power of individual nogoods by resolving quantified clauses instead of their ground instances.

Finally, we remark that a representation very similar to QPROP has also been used in Answer Set Programming (ASP) (Marek & Truszczynski, 1999; Niemelä, 1999) under the name "propositional schemata" (East & Truszczyński, 2001, 2002). The approach used in ASP resembles existing satisfiability work, however, in that clauses are always grounded out. The potential advantages of intelligent subsearch are thus not exploited, although we expect that many of the motivations and results given here would also apply in ASP. In fact, ASP has many features in common with SAT:

- In the most commonly used semantics, that of (non-disjunctive) stable model logic programming (Gelfond & Lifschitz, 1988), the representational power is precisely that of NP (or $NP^{NP}$ for disjunctive programming).

- Cardinality constraints are allowed (East & Truszczyński, 2002; Simons, 2000).

- Solution methods (Leone, Pfeifer, & et al., 2002; Niemelä, 1999; Simons, 2000) use DPLL and some form of propagation.

The most significant difference between conventional satisfiability work and ASP with stable model semantics is that the relevant logic is not classical but the "logic of here and there" (Pearce, 1997). In the logic of here and there, the law of the excluded middle does not hold, only the weaker $\neg p \lor \neg\neg p$. This is sufficient for DPLL to be applied, but does imply that classical resolution is no longer valid. As a result, there seems to be no proof theory for the resulting system, and learning within this framework is not yet understood. Backjumping is used, but the mechanism does not seem to learn new rules from failed subtrees in the search. In an analogous way, cardinality constraints are used but cutting plane proof systems are not. Despite the many parallels between SAT and ASP, including the approach in this survey seems to be somewhat premature.

## 6. Conclusion

Satisfiability algorithms have too often been developed against the framework provided by either random instances or, worse still, instances that have been designed solely to show that the technique being proposed has computational merit. The algorithms themselves have thus tended to ignore problem features that dominate the computational requirements when they are applied to real problems.

On such realistic problems, it is possible to both improve the speed of the algorithms' inner loops (via QPROP and subsearch) and to reduce the number of times that the inner loops need to be executed (via learning and a move up the $p$-simulation hierarchy). Both of these classes of improvements arise because the problems in question have *structure*. The structure can be learned as nogoods, or used to re-represent the problem using pseudo-Boolean or quantified expressions.

It is true that the table in the previous subsection can be viewed as a survey of recent work on satisfiability, and it is also true that the table can be viewed as a rational reconstruction of the goals of the researchers who have investigated various representational extensions. But to our mind, the table is more accurately viewed as a report on the extent to which these linguistic or semantic modifications successfully capture problem structure.

Every column in the table is about structure. Improved representational efficiency is only possible if the problem itself has structure that a Boolean axiomatization typically obscures. It is structure that allows progress to be made in terms of proof complexity. The structure must be preserved by the basic inference mechanism of the system in question if it is to remain useful, and QPROP's ability to speed the inner loop of unit propagation is a direct consequence of the structure present in the subsearch problem. Finally, learning itself can be thought of as a search for reasonably concise descriptions of large sections of the search space that contain no solutions – in other words, learning is the discovery of structure in the search space itself.

This is the setting against which the next two papers in this series are set. If so much of the progress in satisfiability techniques can be thought of as structure exploitation, then surely it is natural to attempt to understand and to exploit that structure directly. As we will see, not only do the techniques we have discussed work by exploiting structure, but they all exploit different instances of a *single* structure. The ZAP work is an attempt to understand, generalize and streamline previous results by setting them in this uniform setting.

## Acknowledgments

We would like to thank the members of CIRL, the technical staff of On Time Systems, and Eugene Luks and David Hofer from the CIS department at the University of Oregon for their assistance with the ideas in this series of papers. We would also like to thank the anonymous reviewers for their comments and suggestions, which we found extremely valuable.

This work was sponsored in part by grants from Air Force Office of Scientific Research (AFOSR) number F49620-92-J-0384, the Air Force Research Laboratory (AFRL) number F30602-97-0294, Small Business Technology Transfer Research, Advanced Technology Institute (STTR-ATI) number 20000766, Office of Naval Research (ONR) number N00014-00-C-0233, and the Defense Advanced Research Projects Agency (DARPA) and the Air Force Research Laboratory, Rome, NY, under agreements numbered F30602-95-1-0023, F30602-97-1-0294, F30602-98-2-0181, F30602-00-2-0534, and F33615-02-C-4032. The views expressed are those of the authors.